\pdfoutput=1

\documentclass[11pt]{article}

\usepackage{EMNLP2022}

\usepackage{times}
\usepackage{latexsym}

\usepackage[T2A, T1]{fontenc}

\usepackage[utf8]{inputenc}

\usepackage{microtype}

\usepackage{inconsolata}

\usepackage{graphicx}
\usepackage{tabularx}
\usepackage{makecell}
\usepackage{booktabs}
\usepackage{multirow}
\usepackage{subcaption}
\usepackage{enumitem}

\definecolor{figuregreen}{rgb}{0.0,0.4,0.0}
\definecolor{figureblue}{rgb}{0.0,0.4,0.7}

\title{Text Characterization Toolkit}

\author{
{\bf Daniel Simig\textsuperscript{*}},
{\bf Tianlu Wang\textsuperscript{*}},
{\bf Verna Dankers\textsuperscript{*+}},
{\bf Peter Henderson\textsuperscript{\ddag*}},
\\
{\bf Khuyagbaatar Batsuren\textsuperscript{\textdagger}},
{\bf Dieuwke Hupkes\textsuperscript{*}},
{\bf Mona Diab\textsuperscript{*}} \\
\textsuperscript{*}Meta AI,
\textsuperscript{+}University of Edinburgh,
\textsuperscript{\textdagger}National University of Mongolia,
\textsuperscript{\ddag}Stanford University \\
\texttt{\{danielsimig,dieuwkehupkes,mdiab\}@fb.com}
}

\begin{document}
\maketitle

\begin{abstract}
In NLP, models are usually evaluated by reporting single-number performance scores on a number of readily available benchmarks, without much deeper analysis.
Here, we argue that -- especially given the well-known fact that benchmarks often contain biases, artefacts, and spurious correlations -- deeper results analysis should become the de-facto standard when presenting new models or benchmarks.
We present a tool that researchers can use to study properties of the dataset and the influence of those properties on their models' behaviour.
Our \emph{Text Characterization Toolkit} includes both an easy-to-use annotation tool, as well as 
off-the-shelf scripts that can be used for specific analyses.
We also
present use-cases from three different domains: we use the tool to predict what are difficult examples for given well-known trained models and identify (potentially harmful) biases and heuristics that are present in a dataset.


\end{abstract}
\section{Introduction}
NLP technology has progressed tremendously over the recent decades with significant advances in algorithms and modeling. Yet, by comparison, our understanding lags behind significantly for datasets (including all datasets types in the model life cycle: training, validation, evaluation) that contribute to model performance. This is mostly due to the lack of frameworks, methods, and tools to draw insights into datasets, especially at scale.

Most NLP models, to date, are evaluated using a relatively small number of readily available evaluation benchmarks, that are often created automatically, or via crowd-sourcing \citep[e.g.][]{bowman-etal-2015-large,wang-etal-2018-glue,williams-etal-2018-broad,zellers-etal-2018-swag}.
It is well-known that most popular (evaluation) datasets are rife with biases, dataset artefacts and spurious correlations, and are prone to be solved with shortcuts \citep{gardner-etal-2021-competency,kiela-etal-2021-dynabench}.
Presenting models with \emph{adversarial examples} for which those biases or correlations do not hold, often results in stark performance drops \citep[e.g.][]{linzen-2020-accelerate,mccoy-etal-2019-right,jia-liang-2017-adversarial,chen-etal-2016-thorough,poliak-etal-2018-hypothesis,tsuchiya-2018-performance,belinkov-etal-2019-dont}. At best, using datasets with such known issues might result in overestimation of a models' capability on the task in question, which may not be reflective of how well they can execute this task in more realistic scenarios.
More worrying, however, is that training or finetuning on datasets that contain biases and artefacts may result in models implementing undesired, biased behavior \citep[e.g.][]{rudinger-etal-2018-gender,blodgett-etal-2016-demographic}.

Additionally,  datasets are usually treated as homogeneous collections of text, performance for which is captured in a single number -- even though there is often a substantial difference between the difficulty/complexity of different examples in a dataset \citep[e.g.][]{sugawara-etal-2022-makes}. Research papers rarely report thorough analysis of performance broken down by characteristics of the data set examples ignoring underlying patterns performance numbers may reflect. The problem is exacerbated by the pervasiveness of benchmarks coupled with a leaderboard competitive culture, where what counts most is system rank.  

In part, this may be due to the fact that deeper analysis of results -- especially when a number of different datasets is involved -- is complex and time-consuming, and there are no standard frameworks or protocols that practitioners can resort to.
The problem is even more pervasive, where we curate datasets for development and evaluation. How we curate, create, select data plays a critical role in understanding our models. Many NLP models (even beyond text) require up/down sampling of specific types of data. These processes should rely on principled characterization of data for any given model. 

Towards this end, we believe that the existence of a standard toolkit that provides an easy to use set of tools and metrics allowing researchers to analyze and systematically characterize  datasets involved in the model life cycle, while gaining insights into the relationship between model performance and data properties could become more common place. 

In this paper, we introduce the \emph{Text Characterization Toolkit}\footnote{\url{https://github.com/facebookresearch/text\_characterization\_toolkit}} (TCT), which aims to enable researchers to gain a detailed understanding of the datasets and models they create -- with minimal effort. TCT is inspired by the Coh-Metrix toolkit \cite{Graesser2004Coh-Metrix:Language}, a collection of over 100 diverse text characteristics intended for use for text analysis in various applications.
TCT offers these capabilities at scale by design. While TCT can process a dataset of 20000 paragraphs in less than a minute using a single command on a MacBook Pro laptop, the very same library, for instance, can also be used as part of a PySpark pipeline to compute text characteristics for a full snapshot of Common Crawl\footnote{\url{https://commoncrawl.org}} (3.1B web pages) in a matter of hours. In this paper we present:
\begin{enumerate}[noitemsep, topsep=1pt]
    \item A repository of text metrics that can help  reveal (hidden) patterns in datasets coupled with model performance on these datasets;
    \item A set of off-the-shelf analysis tools that researchers can use in a simple notebook to study properties of the dataset and the influence of those properties on model behaviour;
    \item A framework that enables the community to share, reuse and standardize metrics and analyses methods/tools;
    \item Use cases that demonstrate the efficacy of TCT in practice covering Language Model prompting, Translation and Bias Detection.
\end{enumerate}

With these contributions, we aspire to contribute to improving how we assess NLP models, and get closer to a scenario where providing detailed results analyses becomes the standard for NLP research.

\section{The Text Characterization Toolkit}
\label{sec:toolkit}

\begin{figure}[t]
\captionsetup{justification=centering}
    \centering
    \includegraphics[width=0.9\linewidth]{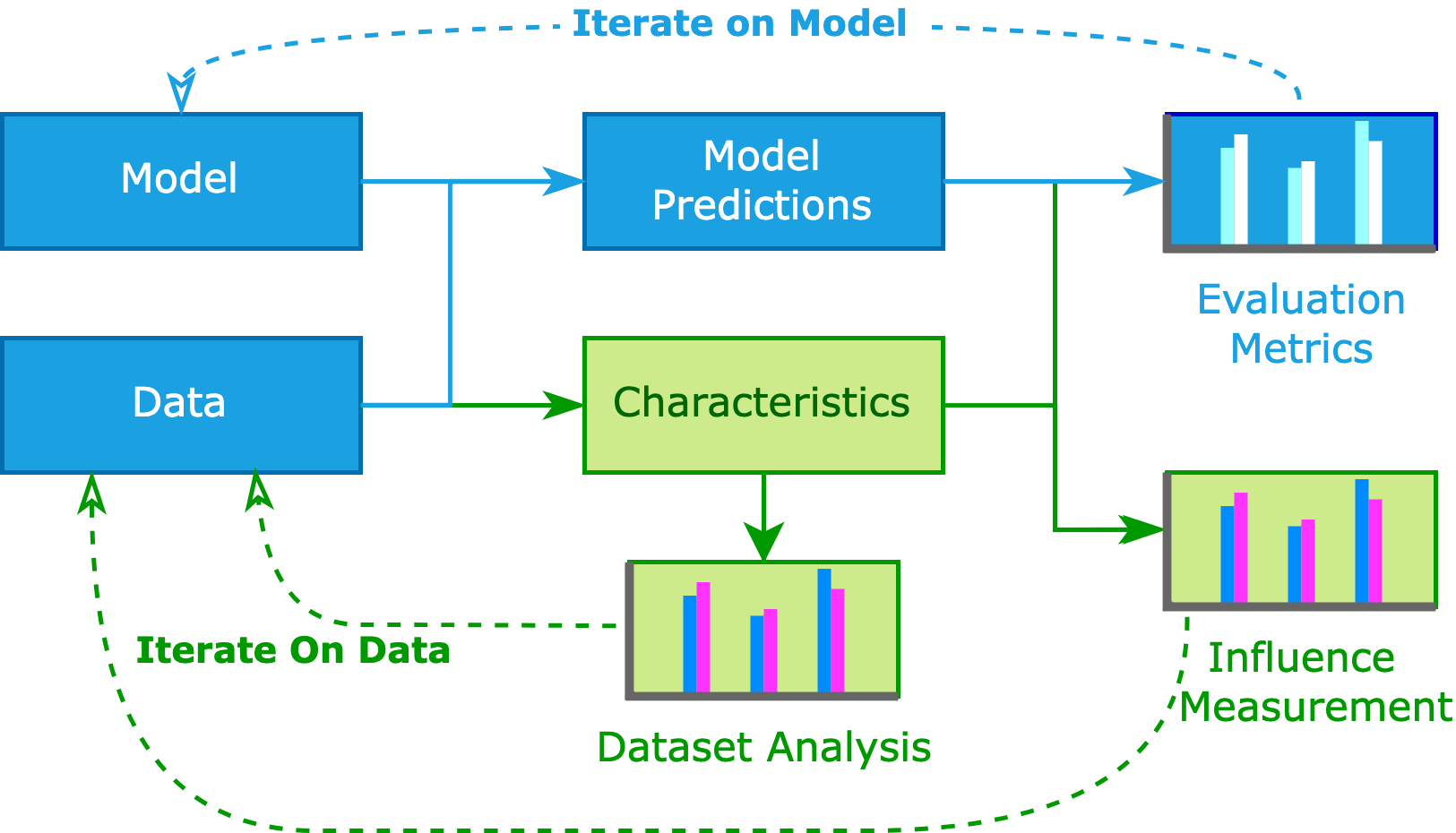}
    \caption{\textcolor{figuregreen}{Text Characterization Toolkit} extends \textcolor{figureblue}{model evaluation} to provide insights about the role of data.}
    \label{fig:overview}
    \vspace{-0.5cm}
\end{figure}

TCT consists of two main components:
\begin{itemize}[noitemsep, topsep=1pt]
    \item A framework for defining and computing text characteristics.
    \item A collection of analysis tools that help users interpret text characteristics and evaluate results with respect to these characteristics. 
\end{itemize}

As illustrated by Figure \ref{fig:overview}, the workflow of extending a standard evaluation process with TCT is typically the following:

\begin{itemize}[noitemsep, topsep=0pt]
    \item Given a dataset, define how to extract text fragments from each data point: For a QA dataset text fragments could be individual questions, whereas in document summarization, the text fragments would be the documents themselves.
    \item Use TCT to compute characteristics of the text fragments. One might use the default characteristics already included in the framework or define their own specific metric.
    \item Load the computed characteristics and other evaluation specific data into a Python notebook for analysis using TCT. One might analyze the dataset itself (e.g. to identify spurious correlations or biases) or jointly analyze model evaluation metrics and text characteristics (e.g. through correlation analysis between TCT features and models' test set accuracy).
    \item Use the results of the analysis  to improve the dataset, the model, or the evaluation protocol -- for example by extending evaluation data with examples where a model is expected to perform poorly or focusing evaluation on a challenging subset of the test data.
\end{itemize}

Concrete examples of the workflow above are described in \S\ref{section:use_cases_short} and in Appendix \ref{section:use_cases}. The rest of this section provides more details on the two important components of the framework.

\subsection{Text Characteristics}

\begin{table}
\captionsetup{justification=centering}
\centering\small
\begin{tabular}{lc}
\toprule
 \textbf{Category} & \textbf{Example Metrics}  \\ 
 \midrule\midrule
 \multirow{2}{*}{Descriptive} & Word Count \\
 & Sentence Length\\  
 \midrule
 \multirow{2}{*}{Lexical Diversity} & Type-Token Ratio \\
 & MTLD \\  
 \midrule
 \multirow{2}{*}{Complexity} & Left Embeddedness \\
 & \# of NP modifiers \\
 \midrule
\multirow{2}{*}{Incidence Scores} & Different POS tags \\
& Types of connectives \\
 \midrule
 \multirow{2}{*}{Word Property} & Age of Acquisition \\
 & Concreteness \\
\bottomrule
\end{tabular}
\caption{Categories of characteristics currently implemented. See Appendix \ref{appendix:metrics} for an exhaustive list.}
\label{fig:metric_summary}
\vspace{-0.5cm}
\end{table}


While the majority of the characteristics found in TCT is motivated by metric classes in Coh-Metrix \cite{Graesser2004Coh-Metrix:Language}, we have included new data bases for existing metrics and added entirely new metrics. At the time of writing, there are 61 characteristics implemented in TCT. An overview of the main categories of currently implemented characteristics can be found in Table \ref{fig:metric_summary}.
The toolkit provides a standardized framework to implement, configure, and compute these metrics. Adding a new metric is as simple as implementing two functions: one that loads any required resource (such as a word database) and initializes computation, and one that computes the metric given these resources and an input text. 


\subsection{Analysis tools}
\begin{figure}
    \captionsetup{justification=centering}
    \begin{subfigure}{.45\linewidth}
        \centering
        \includegraphics[width=\linewidth]{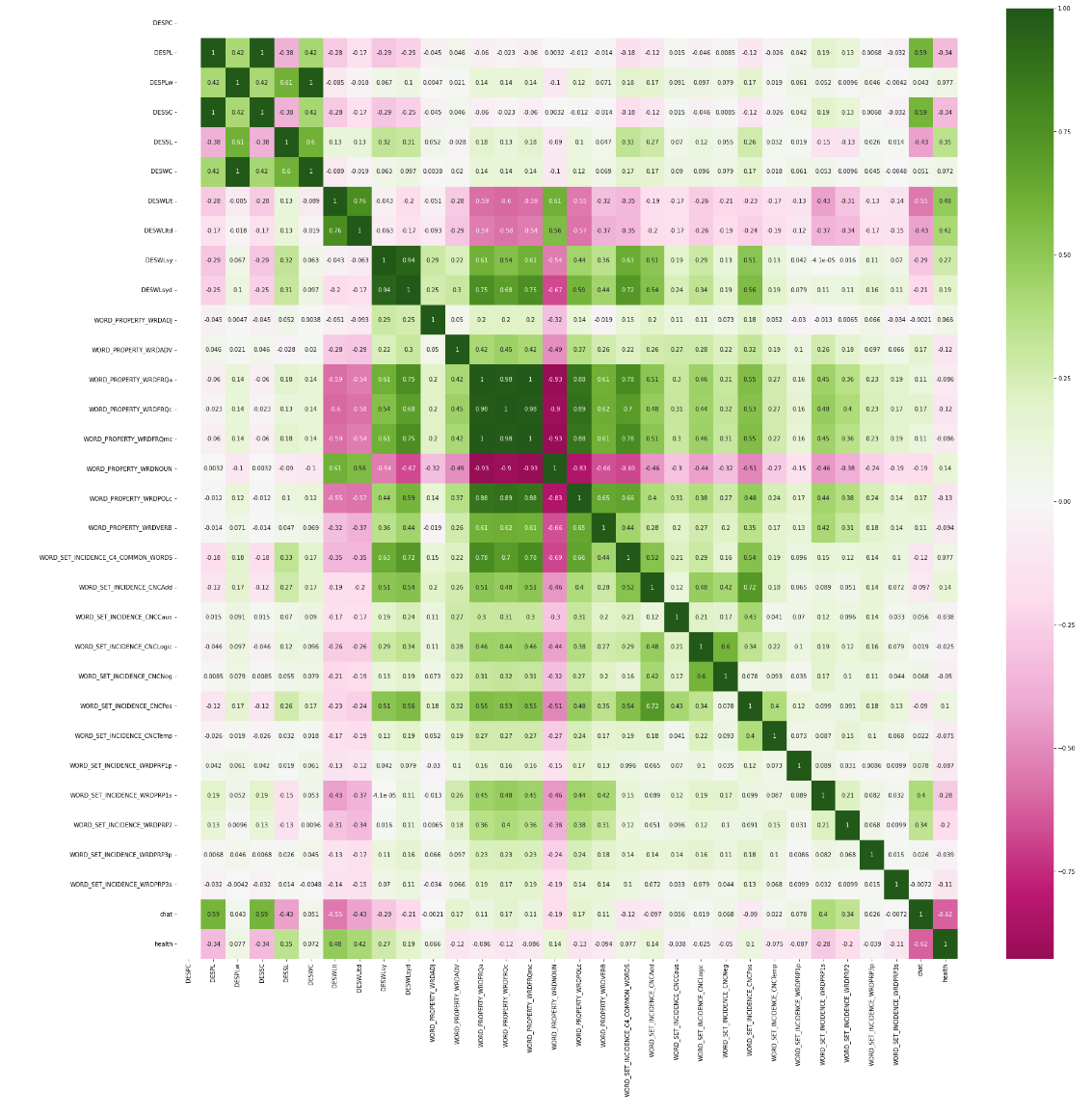}
        \caption{Correlations between text characteristics}
        \label{fig:sub1}
    \end{subfigure}%
    \hfill
    \begin{subfigure}{.45\linewidth}
        \centering
        \includegraphics[width=\linewidth]{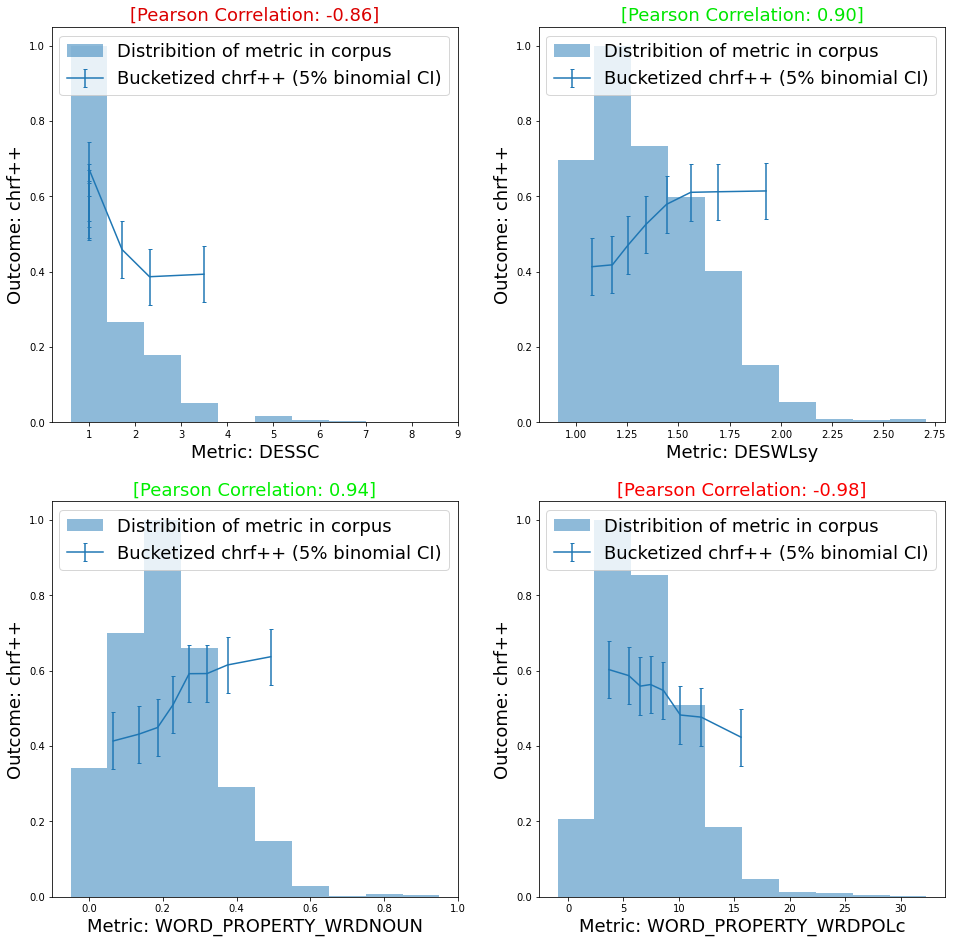}
        \caption{Model performance w.r.t. some characteristics}
        \label{fig:sub2}
    \end{subfigure}\\[1ex]
    \begin{subfigure}{\linewidth}
        \centering
        \raisebox{-0.5\height}{\includegraphics[width=0.3\linewidth]{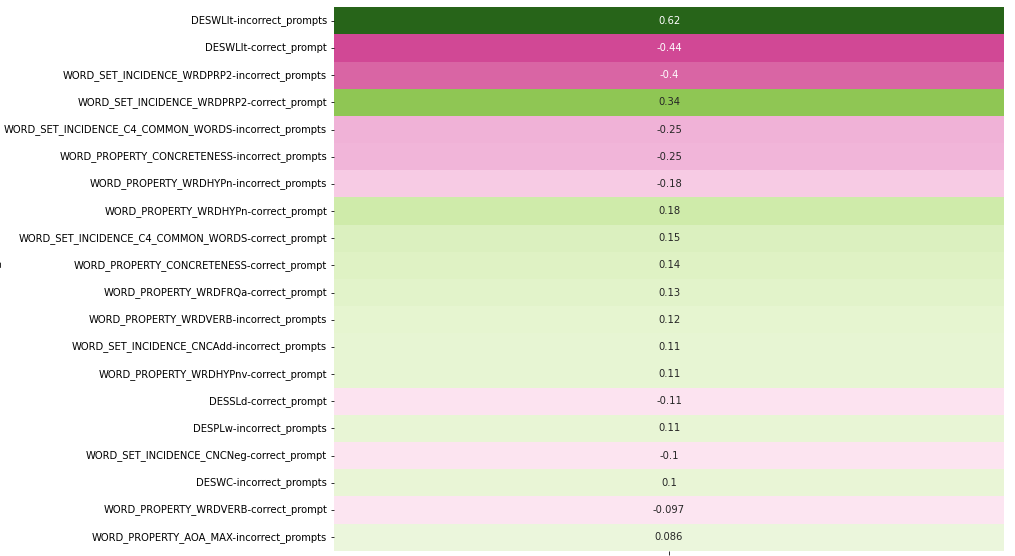}}
        \raisebox{-0.5\height}{\includegraphics[width=0.3\linewidth]{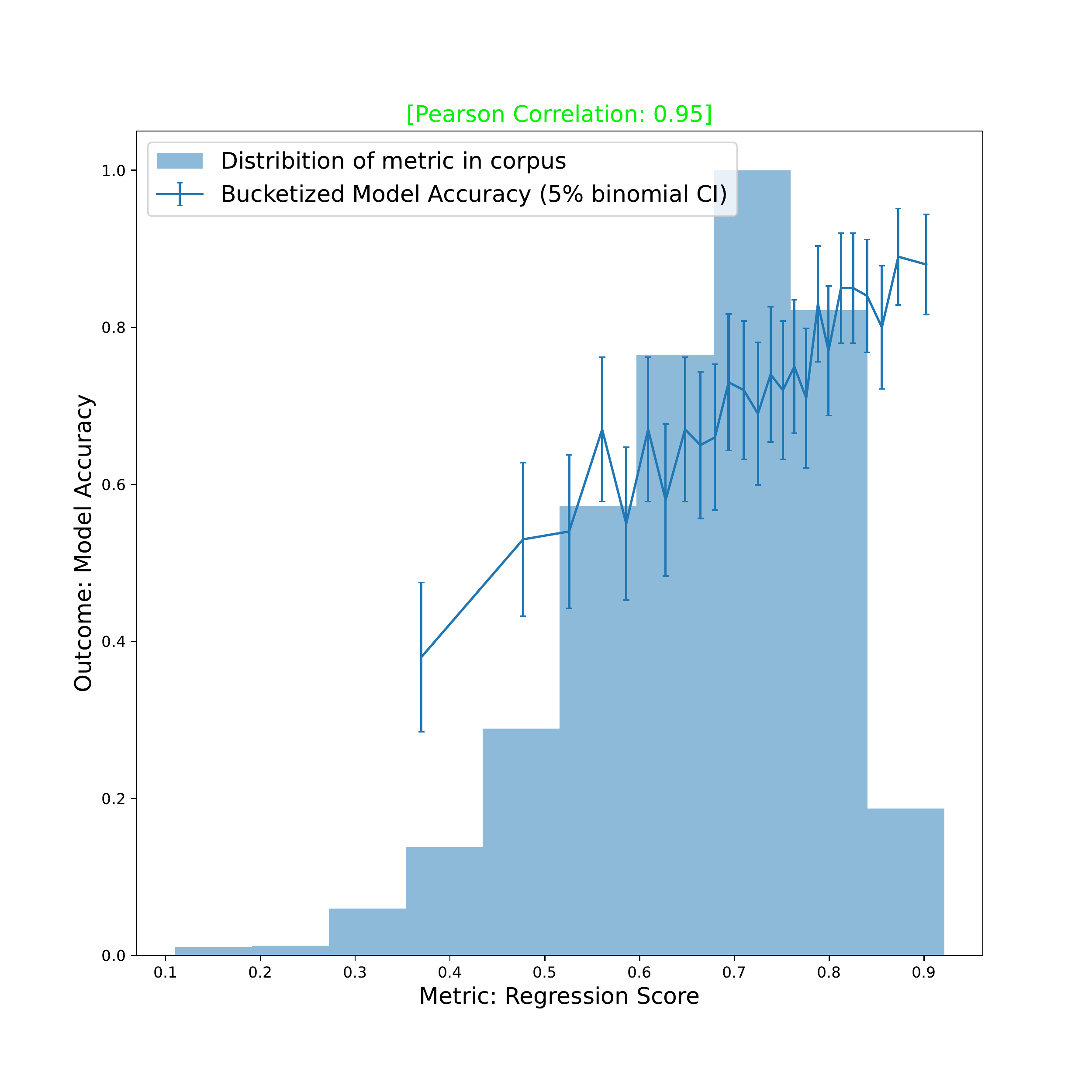}}
        \caption{Results of a regression analysis: coefficients and fit}
        \label{fig:sub3}
    \end{subfigure}
    \caption{TCT analysis tools in action. See Appendix \ref{section:use_cases} for detailed explanations and high-resolution images.}
    \label{fig:tools_overview}
\end{figure}

To further decrease the effort required to carry out text characteristics based analysis, we provide an initial set of analysis tools that users can use out of the box. We encourage users to contribute their own implementations of TCT-based analyses to the toolkit, to allow for re-use in the future development of datasets and models. The current functionality of the toolkit, as illustrated in Figure \ref{fig:tools_overview}, includes:
\begin{enumerate}[noitemsep, topsep=1pt]
    \item Visualising distributions of different characteristics;
    \item Visualising a pairwise correlation matrix for the characteristics;
    \item Visualising correlations between individual characteristics and outcomes (e.g., accuracy);
    \item Fitting a model on all characteristics to outcomes (logistic regression and random forests are supported currently) and analyzing a model’s predictive power and coefficients.
\end{enumerate}

\section{Example Use Cases}
\label{section:use_cases_short}

In order to demonstrate the ability of TCT to produce meaningful and actionable insights, we provide 3 examples of its use on real world data. For each one of these use cases, a thorough description of the experimental setup and results is included in Appendix \ref{section:use_cases} and reference notebooks are provided in the \texttt{examples} directory of the TCT repository.

\paragraph{Predicting Accuracy of OPT Baselines}
We use the logistic regression analysis tool to fit a model that predicts the accuracy of the 6.7B OPT \cite{Zhang2022OPT:Models}  model on the HellaSwag \cite{Zellers2019HellaSwag:Sentence} task based on simple characteristics such as mean word length and concreteness. Using this model we identify subsets of the test data with precision as low as 40\% and as high as 90\%.

\paragraph{Gender Bias in Co-reference Resolution}
By computing genderedness metrics on co-reference labels and using these metrics as inputs to the analysis tools, we reproduce the results of ~\citet{zhao-etal-2018-gender} showing that models perform much worse when the stereotypically associated gender of an occupation does not match the gender of the pronominal reference.

\paragraph{Fluctuations in Translation Performance}

We show how translation performance of the NLLB model~\citep{costa2022no} using the HuggingFace pipeline~\citep{wolf2019huggingface} fluctuates as a function of sample characteristics, like the number of sentences. This performance heterogeneity can be fixed by segmenting sentences before using the pipeline, showing that TCT can help debug model pipelines even with many layers of abstraction.
\section{Related Work}

Multiple existing tools offer similar functionality as TCT does:
DataLab \cite{Xiao2022DataLab:Intervention} is a tool for detailed data analysis that, among other things, allows users to inspect datasets through the lens of a few text characteristics such as text length, lexical diversity and gender-related features. The \textit{Know Your Data}\footnote{\url{https://knowyourdata.withgoogle.com/}} tool allows for inspection of image data, it surfaces spurious correlations, biases and imbalances in datasets. However, both tools do not connect model behavior to properties of datasets.
~\citet{collins-etal-2018-evolutionary} predicts overall hardness of classification datasets based on label statistics and a few text characteristics such as readability and lexical diversity. ExplainaBoard \cite{Liu2021ExplainaBoard:NLP} focuses on model performance analysis and provides a model performance breakdown by simple attributes of data points such as text length, providing a functionality most similar to our work.

Our toolkit distinguishes itself by including a much wider range of text characteristics and multi-variable analysis tools that can identify larger variations in model accuracy. 
By packaging our toolkit as a simple Python library used in notebooks -- in contrast to the previously described feature-rich systems -- we also intend to minimize the effort needed to both use it as well as contribute to it (crowd sourcing more functionality).

The Coh-Metrix tool \cite{Graesser2004Coh-Metrix:Language} collected the most diverse set of text characteristics to our knowledge, designed for various use cases in linguistics and pedagogy. 
The tool, developed in 2004, is slow as it is designed to process a single document at a time, relatively difficult to access, and the underlying word databases are outdated. 
Our toolkit aims to make a subset of these metrics easily accessible to the NLP community.

\section{Future Work}

As illustrated in \S\ref{sec:toolkit} we envision TCT to be a framework and an associated tool that allows for community contributions, crowdsourcing even more functionality and use cases. Future work involves usage of the tool:

Firstly, we encourage creators of new datasets to use TCT as a data annotation tool, to extract a wide range of dataset statistics in a straightforward manner, and report about them in academic publications for transparency about the contents of their dataset.
Such statistics could be included in  datasheets and data cards \citep{gebru2021datasheets}, and they can aid in outlier detection during data cleaning.

We also prompt dataset creators to perform statistical analyses capturing which features are predictive of the gold targets \textit{before} further training computational models, to ensure one is aware about potential short-cut learning opportunities due to biases in the dataset. Naturally, not all correlations are bad or avoidable -- e.g.\ consider sentences containing the word `fantastic' that are likely to have a positive label in sentiment analysis -- but others are good to be aware of when working with a dataset -- e.g.\ consider a natural language inference task where all sentences with the label `entailed' have an atypical average word length. Such analyses could be included in a `cautions' section with a dataset's release.

A third type of usage would be by owners of new models, that, on the one hand, use TCT to measure whether some dataset characteristics are predictive of success and failure by their model, and, on the other hand, provide performance on subclasses of samples. One may already know that model performance is lower for longer sentences, but what about performance on different readability classes, classes with varying amounts of causal connectives or different ratings for syntactic complexity (e.g. SYNLE)? TCT will help answer those questions. Understanding how the model performance fluctuates for different data subsets provides further understanding in model robustness, and can, in turn, improve datasets' quality if model owners report back on biases identified in datasets.
It should be noted that TCT could be an effective tool for data selection for both training and evaluation, in particular at scale.

\section*{Limitations}
Text Characteristics in our framework have varying levels of coverage depending on their type. Word property based characteristics, for example, are limited by the coverage of the word databases that back them -- this can be limited even for English. While we plan to extend the framework to multiple languages in the near future, language coverage of backing word databases and NLP pipelines such as WordNet \cite{10.1145/219717.219748} or SpaCy \cite{Honnibal_spaCy_Industrial-strength_Natural_2020} will affect our ability to scale the number of languages supported.

\bibliography{anthology,custom,daniel_mendeley}

\begin{thebibliography}{39}
\expandafter\ifx\csname natexlab\endcsname\relax\def\natexlab#1{#1}\fi

\bibitem[{Belinkov et~al.(2019)Belinkov, Poliak, Shieber, Van~Durme, and
  Rush}]{belinkov-etal-2019-dont}
Yonatan Belinkov, Adam Poliak, Stuart Shieber, Benjamin Van~Durme, and
  Alexander Rush. 2019.
\newblock \href {https://doi.org/10.18653/v1/P19-1084} {Don{'}t take the
  premise for granted: Mitigating artifacts in natural language inference}.
\newblock In \emph{Proceedings of the 57th Annual Meeting of the Association
  for Computational Linguistics}, pages 877--891, Florence, Italy. Association
  for Computational Linguistics.

\bibitem[{Blodgett et~al.(2016)Blodgett, Green, and
  O{'}Connor}]{blodgett-etal-2016-demographic}
Su~Lin Blodgett, Lisa Green, and Brendan O{'}Connor. 2016.
\newblock \href {https://doi.org/10.18653/v1/D16-1120} {Demographic dialectal
  variation in social media: A case study of {A}frican-{A}merican {E}nglish}.
\newblock In \emph{Proceedings of the 2016 Conference on Empirical Methods in
  Natural Language Processing}, pages 1119--1130, Austin, Texas. Association
  for Computational Linguistics.

\bibitem[{Bowman et~al.(2015)Bowman, Angeli, Potts, and
  Manning}]{bowman-etal-2015-large}
Samuel~R. Bowman, Gabor Angeli, Christopher Potts, and Christopher~D. Manning.
  2015.
\newblock \href {https://doi.org/10.18653/v1/D15-1075} {A large annotated
  corpus for learning natural language inference}.
\newblock In \emph{Proceedings of the 2015 Conference on Empirical Methods in
  Natural Language Processing}, pages 632--642, Lisbon, Portugal. Association
  for Computational Linguistics.

\bibitem[{Brysbaert et~al.()Brysbaert, Mandera, Mccormick, and
  Keuleers}]{BrysbaertWordLemmas}
Marc Brysbaert, Paweł Mandera, Samantha~F Mccormick, and Emmanuel Keuleers.
\newblock \href {https://doi.org/10.3758/s13428-018-1077-9} {{Word prevalence
  norms for 62,000 English lemmas}}.

\bibitem[{Brysbaert et~al.(2014)Brysbaert, Warriner, and
  Kuperman}]{Brysbaert2014ConcretenessLemmas}
Marc Brysbaert, Amy~Beth Warriner, and Victor Kuperman. 2014.
\newblock \href {https://doi.org/10.3758/S13428-013-0403-5/FIGURES/1}
  {{Concreteness ratings for 40 thousand generally known English word lemmas}}.
\newblock \emph{Behavior Research Methods}, 46(3):904--911.

\bibitem[{Chen et~al.(2016)Chen, Bolton, and Manning}]{chen-etal-2016-thorough}
Danqi Chen, Jason Bolton, and Christopher~D. Manning. 2016.
\newblock \href {https://doi.org/10.18653/v1/P16-1223} {A thorough examination
  of the {CNN}/{D}aily {M}ail reading comprehension task}.
\newblock In \emph{Proceedings of the 54th Annual Meeting of the Association
  for Computational Linguistics (Volume 1: Long Papers)}, pages 2358--2367,
  Berlin, Germany. Association for Computational Linguistics.

\bibitem[{Chowdhery et~al.(2022)Chowdhery, Narang, Devlin, Bosma, Mishra,
  Roberts, Barham, Chung, Sutton, Gehrmann et~al.}]{chowdhery2022palm}
Aakanksha Chowdhery, Sharan Narang, Jacob Devlin, Maarten Bosma, Gaurav Mishra,
  Adam Roberts, Paul Barham, Hyung~Won Chung, Charles Sutton, Sebastian
  Gehrmann, et~al. 2022.
\newblock Palm: Scaling language modeling with pathways.
\newblock \emph{arXiv e-prints}, pages arXiv--2204.

\bibitem[{Collins et~al.(2018)Collins, Rozanov, and
  Zhang}]{collins-etal-2018-evolutionary}
Edward Collins, Nikolai Rozanov, and Bingbing Zhang. 2018.
\newblock \href {https://doi.org/10.18653/v1/K18-1037} {Evolutionary data
  measures: Understanding the difficulty of text classification tasks}.
\newblock In \emph{Proceedings of the 22nd Conference on Computational Natural
  Language Learning}, pages 380--391, Brussels, Belgium. Association for
  Computational Linguistics.

\bibitem[{Coltheart(2018)}]{Coltheart2018TheDatabase:}
Max Coltheart. 2018.
\newblock \href {https://doi.org/10.1080/14640748108400805} {{The MRC
  Psycholinguistic Database:}}.
\newblock \emph{https://doi.org/10.1080/14640748108400805}, 33(4):497--505.

\bibitem[{Costa-juss{\`a} et~al.(2022)Costa-juss{\`a}, Cross, {\c{C}}elebi,
  Elbayad, Heafield, Heffernan, Kalbassi, Lam, Licht, Maillard
  et~al.}]{costa2022no}
Marta~R Costa-juss{\`a}, James Cross, Onur {\c{C}}elebi, Maha Elbayad, Kenneth
  Heafield, Kevin Heffernan, Elahe Kalbassi, Janice Lam, Daniel Licht, Jean
  Maillard, et~al. 2022.
\newblock No language left behind: Scaling human-centered machine translation.
\newblock \emph{arXiv e-prints}, pages arXiv--2207.

\bibitem[{Fellbaum(2010)}]{Fellbaum2010WordNet}
Christiane Fellbaum. 2010.
\newblock \href {https://doi.org/10.1007/978-90-481-8847-5{\_}10/COVER}
  {{WordNet}}.
\newblock \emph{Theory and Applications of Ontology: Computer Applications},
  pages 231--243.

\bibitem[{Gardner et~al.(2021)Gardner, Merrill, Dodge, Peters, Ross, Singh, and
  Smith}]{gardner-etal-2021-competency}
Matt Gardner, William Merrill, Jesse Dodge, Matthew Peters, Alexis Ross, Sameer
  Singh, and Noah~A. Smith. 2021.
\newblock \href {https://doi.org/10.18653/v1/2021.emnlp-main.135} {Competency
  problems: On finding and removing artifacts in language data}.
\newblock In \emph{Proceedings of the 2021 Conference on Empirical Methods in
  Natural Language Processing}, pages 1801--1813, Online and Punta Cana,
  Dominican Republic. Association for Computational Linguistics.

\bibitem[{Gebru et~al.(2021)Gebru, Morgenstern, Vecchione, Vaughan, Wallach,
  Iii, and Crawford}]{gebru2021datasheets}
Timnit Gebru, Jamie Morgenstern, Briana Vecchione, Jennifer~Wortman Vaughan,
  Hanna Wallach, Hal~Daum{\'e} Iii, and Kate Crawford. 2021.
\newblock Datasheets for datasets.
\newblock \emph{Communications of the ACM}, 64(12):86--92.

\bibitem[{Graesser et~al.(2004)Graesser, McNamara, Louwerse, and
  Cai}]{Graesser2004Coh-Metrix:Language}
Arthur~C. Graesser, Danielle~S. McNamara, Max~M. Louwerse, and Zhiqiang Cai.
  2004.
\newblock \href {https://doi.org/10.3758/BF03195564} {{Coh-Metrix: Analysis of
  text on cohesion and language}}.
\newblock \emph{Behavior Research Methods, Instruments, {\&} Computers 2004
  36:2}, 36(2):193--202.

\bibitem[{He et~al.(2020)He, Liu, Gao, and Chen}]{he2020deberta}
Pengcheng He, Xiaodong Liu, Jianfeng Gao, and Weizhu Chen. 2020.
\newblock {DeBERTa}: decoding-enhanced {BERT} with disentangled attention.
\newblock In \emph{International Conference on Learning Representations}.

\bibitem[{Honnibal et~al.(2020)Honnibal, Montani, Van~Landeghem, and
  Boyd}]{Honnibal_spaCy_Industrial-strength_Natural_2020}
Matthew Honnibal, Ines Montani, Sofie Van~Landeghem, and Adriane Boyd. 2020.
\newblock \href {https://doi.org/10.5281/zenodo.1212303} {{spaCy:
  Industrial-strength Natural Language Processing in Python}}.

\bibitem[{Jia and Liang(2017)}]{jia-liang-2017-adversarial}
Robin Jia and Percy Liang. 2017.
\newblock \href {https://doi.org/10.18653/v1/D17-1215} {Adversarial examples
  for evaluating reading comprehension systems}.
\newblock In \emph{Proceedings of the 2017 Conference on Empirical Methods in
  Natural Language Processing}, pages 2021--2031, Copenhagen, Denmark.
  Association for Computational Linguistics.

\bibitem[{Kiela et~al.(2021)Kiela, Bartolo, Nie, Kaushik, Geiger, Wu, Vidgen,
  Prasad, Singh, Ringshia, Ma, Thrush, Riedel, Waseem, Stenetorp, Jia, Bansal,
  Potts, and Williams}]{kiela-etal-2021-dynabench}
Douwe Kiela, Max Bartolo, Yixin Nie, Divyansh Kaushik, Atticus Geiger,
  Zhengxuan Wu, Bertie Vidgen, Grusha Prasad, Amanpreet Singh, Pratik Ringshia,
  Zhiyi Ma, Tristan Thrush, Sebastian Riedel, Zeerak Waseem, Pontus Stenetorp,
  Robin Jia, Mohit Bansal, Christopher Potts, and Adina Williams. 2021.
\newblock \href {https://doi.org/10.18653/v1/2021.naacl-main.324} {Dynabench:
  Rethinking benchmarking in {NLP}}.
\newblock In \emph{Proceedings of the 2021 Conference of the North American
  Chapter of the Association for Computational Linguistics: Human Language
  Technologies}, pages 4110--4124, Online. Association for Computational
  Linguistics.

\bibitem[{Kuperman et~al.()Kuperman, Stadthagen-Gonzalez, and
  Brysbaert}]{KupermanAge-of-acquisitionWords}
Victor Kuperman, Hans Stadthagen-Gonzalez, and Marc Brysbaert.
\newblock \href {https://doi.org/10.3758/s13428-012-0210-4}
  {{Age-of-acquisition ratings for 30,000 English words}}.

\bibitem[{Lee et~al.(2017)Lee, He, Lewis, and Zettlemoyer}]{lee-etal-2017-end}
Kenton Lee, Luheng He, Mike Lewis, and Luke Zettlemoyer. 2017.
\newblock \href {https://doi.org/10.18653/v1/D17-1018} {End-to-end neural
  coreference resolution}.
\newblock In \emph{Proceedings of the 2017 Conference on Empirical Methods in
  Natural Language Processing}, pages 188--197, Copenhagen, Denmark.
  Association for Computational Linguistics.

\bibitem[{Linzen(2020)}]{linzen-2020-accelerate}
Tal Linzen. 2020.
\newblock \href {https://doi.org/10.18653/v1/2020.acl-main.465} {How can we
  accelerate progress towards human-like linguistic generalization?}
\newblock In \emph{Proceedings of the 58th Annual Meeting of the Association
  for Computational Linguistics}, pages 5210--5217, Online. Association for
  Computational Linguistics.

\bibitem[{Liu et~al.(2021)Liu, Fu, Xiao, Yuan, Chang, Dai, Liu, Ye, and
  Neubig}]{Liu2021ExplainaBoard:NLP}
Pengfei Liu, Jinlan Fu, Yang Xiao, Weizhe Yuan, Shuaichen Chang, Junqi Dai,
  Yixin Liu, Zihuiwen Ye, and Graham Neubig. 2021.
\newblock \href {https://doi.org/10.48550/arxiv.2104.06387} {{ExplainaBoard: An
  Explainable Leaderboard for NLP}}.
\newblock \emph{ACL-IJCNLP 2021 - 59th Annual Meeting of the Association for
  Computational Linguistics and the 11th International Joint Conference on
  Natural Language Processing, Proceedings of the System Demonstrations}, pages
  280--289.

\bibitem[{McCoy et~al.(2019)McCoy, Pavlick, and Linzen}]{mccoy-etal-2019-right}
Tom McCoy, Ellie Pavlick, and Tal Linzen. 2019.
\newblock \href {https://doi.org/10.18653/v1/P19-1334} {Right for the wrong
  reasons: Diagnosing syntactic heuristics in natural language inference}.
\newblock In \emph{Proceedings of the 57th Annual Meeting of the Association
  for Computational Linguistics}, pages 3428--3448, Florence, Italy.
  Association for Computational Linguistics.

\bibitem[{Miller(1995)}]{10.1145/219717.219748}
George~A. Miller. 1995.
\newblock \href {https://doi.org/10.1145/219717.219748} {Wordnet: A lexical
  database for english}.
\newblock \emph{Commun. ACM}, 38(11):39–41.

\bibitem[{Papineni et~al.(2002)Papineni, Roukos, Ward, and
  Zhu}]{papineni2002bleu}
Kishore Papineni, Salim Roukos, Todd Ward, and Wei-Jing Zhu. 2002.
\newblock Bleu: a method for automatic evaluation of machine translation.
\newblock In \emph{Proceedings of the 40th annual meeting of the Association
  for Computational Linguistics}, pages 311--318.

\bibitem[{Peter et~al.()Peter, Robert, Richard, Brad, and
  J}]{PeterKincaidRobertFishburneJrRichardLRogersBradSChissomDerivationPersonnelCustom}
Kincaid Peter, Fishburne~Jr Robert, L~Rogers Richard, S~Chissom Brad, and P~J.
\newblock \href {http://library.ucf.edu} {{Derivation Of New Readability
  Formulas (Automated Readability Index, Fog Count And Flesch Reading Ease
  Formula) For Navy Enlisted Personnel}}.

\bibitem[{Poliak et~al.(2018)Poliak, Naradowsky, Haldar, Rudinger, and
  Van~Durme}]{poliak-etal-2018-hypothesis}
Adam Poliak, Jason Naradowsky, Aparajita Haldar, Rachel Rudinger, and Benjamin
  Van~Durme. 2018.
\newblock \href {https://doi.org/10.18653/v1/S18-2023} {Hypothesis only
  baselines in natural language inference}.
\newblock In \emph{Proceedings of the Seventh Joint Conference on Lexical and
  Computational Semantics}, pages 180--191, New Orleans, Louisiana. Association
  for Computational Linguistics.

\bibitem[{Popovi{\'c}(2017)}]{popovic2017chrf++}
Maja Popovi{\'c}. 2017.
\newblock chrf++: words helping character n-grams.
\newblock In \emph{Proceedings of the second conference on machine
  translation}, pages 612--618.

\bibitem[{Rudinger et~al.(2018)Rudinger, Naradowsky, Leonard, and
  Van~Durme}]{rudinger-etal-2018-gender}
Rachel Rudinger, Jason Naradowsky, Brian Leonard, and Benjamin Van~Durme. 2018.
\newblock \href {https://doi.org/10.18653/v1/N18-2002} {Gender bias in
  coreference resolution}.
\newblock In \emph{Proceedings of the 2018 Conference of the North {A}merican
  Chapter of the Association for Computational Linguistics: Human Language
  Technologies, Volume 2 (Short Papers)}, pages 8--14, New Orleans, Louisiana.
  Association for Computational Linguistics.

\bibitem[{Sugawara et~al.(2022)Sugawara, Nangia, Warstadt, and
  Bowman}]{sugawara-etal-2022-makes}
Saku Sugawara, Nikita Nangia, Alex Warstadt, and Samuel Bowman. 2022.
\newblock \href {https://doi.org/10.18653/v1/2022.acl-long.479} {What makes
  reading comprehension questions difficult?}
\newblock In \emph{Proceedings of the 60th Annual Meeting of the Association
  for Computational Linguistics (Volume 1: Long Papers)}, pages 6951--6971,
  Dublin, Ireland. Association for Computational Linguistics.

\bibitem[{Tsuchiya(2018)}]{tsuchiya-2018-performance}
Masatoshi Tsuchiya. 2018.
\newblock \href {https://aclanthology.org/L18-1239} {Performance impact caused
  by hidden bias of training data for recognizing textual entailment}.
\newblock In \emph{Proceedings of the Eleventh International Conference on
  Language Resources and Evaluation ({LREC} 2018)}, Miyazaki, Japan. European
  Language Resources Association (ELRA).

\bibitem[{Wang et~al.(2018)Wang, Singh, Michael, Hill, Levy, and
  Bowman}]{wang-etal-2018-glue}
Alex Wang, Amanpreet Singh, Julian Michael, Felix Hill, Omer Levy, and Samuel
  Bowman. 2018.
\newblock \href {https://doi.org/10.18653/v1/W18-5446} {{GLUE}: A multi-task
  benchmark and analysis platform for natural language understanding}.
\newblock In \emph{Proceedings of the 2018 {EMNLP} Workshop {B}lackbox{NLP}:
  Analyzing and Interpreting Neural Networks for {NLP}}, pages 353--355,
  Brussels, Belgium. Association for Computational Linguistics.

\bibitem[{Williams et~al.(2018)Williams, Nangia, and
  Bowman}]{williams-etal-2018-broad}
Adina Williams, Nikita Nangia, and Samuel Bowman. 2018.
\newblock \href {https://doi.org/10.18653/v1/N18-1101} {A broad-coverage
  challenge corpus for sentence understanding through inference}.
\newblock In \emph{Proceedings of the 2018 Conference of the North {A}merican
  Chapter of the Association for Computational Linguistics: Human Language
  Technologies, Volume 1 (Long Papers)}, pages 1112--1122, New Orleans,
  Louisiana. Association for Computational Linguistics.

\bibitem[{Wolf et~al.(2019)Wolf, Debut, Sanh, Chaumond, Delangue, Moi, Cistac,
  Rault, Louf, Funtowicz et~al.}]{wolf2019huggingface}
Thomas Wolf, Lysandre Debut, Victor Sanh, Julien Chaumond, Clement Delangue,
  Anthony Moi, Pierric Cistac, Tim Rault, R{\'e}mi Louf, Morgan Funtowicz,
  et~al. 2019.
\newblock Huggingface's transformers: State-of-the-art natural language
  processing.
\newblock \emph{arXiv preprint arXiv:1910.03771}.

\bibitem[{Xiao et~al.(2022)Xiao, Fu, Yuan, Viswanathan, Liu, Liu, Neubig, and
  Liu}]{Xiao2022DataLab:Intervention}
Yang Xiao, Jinlan Fu, Weizhe Yuan, Vijay Viswanathan, Zhoumianze Liu, Yixin
  Liu, Graham Neubig, and Pengfei Liu. 2022.
\newblock \href {https://doi.org/10.48550/arxiv.2202.12875} {{DataLab: A
  Platform for Data Analysis and Intervention}}.
\newblock pages 182--195.

\bibitem[{Zellers et~al.(2018)Zellers, Bisk, Schwartz, and
  Choi}]{zellers-etal-2018-swag}
Rowan Zellers, Yonatan Bisk, Roy Schwartz, and Yejin Choi. 2018.
\newblock \href {https://doi.org/10.18653/v1/D18-1009} {{SWAG}: A large-scale
  adversarial dataset for grounded commonsense inference}.
\newblock In \emph{Proceedings of the 2018 Conference on Empirical Methods in
  Natural Language Processing}, pages 93--104, Brussels, Belgium. Association
  for Computational Linguistics.

\bibitem[{Zellers et~al.(2019)Zellers, Holtzman, Bisk, Farhadi, and
  Choi}]{Zellers2019HellaSwag:Sentence}
Rowan Zellers, Ari Holtzman, Yonatan Bisk, Ali Farhadi, and Yejin Choi. 2019.
\newblock \href {https://doi.org/10.48550/arxiv.1905.07830} {{HellaSwag: Can a
  Machine Really Finish Your Sentence?}}
\newblock \emph{ACL 2019 - 57th Annual Meeting of the Association for
  Computational Linguistics, Proceedings of the Conference}, pages 4791--4800.

\bibitem[{Zhang et~al.(2022)Zhang, Roller, Goyal, Artetxe, Chen, Chen, Dewan,
  Diab, Li, Lin, Mihaylov, Ott, Shleifer, Shuster, Simig, Koura, Sridhar, Wang,
  Zettlemoyer, and Ai}]{Zhang2022OPT:Models}
Susan Zhang, Stephen Roller, Naman Goyal, Mikel Artetxe, Moya Chen, Shuohui
  Chen, Christopher Dewan, Mona Diab, Xian Li, Victoria Lin, Todor Mihaylov,
  Myle Ott, Sam Shleifer, Kurt Shuster, Daniel Simig, Singh Koura, Anjali
  Sridhar, Tianlu Wang, Luke Zettlemoyer, and Meta Ai. 2022.
\newblock \href {https://doi.org/10.48550/arxiv.2205.01068} {{OPT: Open
  Pre-trained Transformer Language Models}}.

\bibitem[{Zhao et~al.(2018)Zhao, Wang, Yatskar, Ordonez, and
  Chang}]{zhao-etal-2018-gender}
Jieyu Zhao, Tianlu Wang, Mark Yatskar, Vicente Ordonez, and Kai-Wei Chang.
  2018.
\newblock \href {https://doi.org/10.18653/v1/N18-2003} {Gender bias in
  coreference resolution: Evaluation and debiasing methods}.
\newblock In \emph{Proceedings of the 2018 Conference of the North {A}merican
  Chapter of the Association for Computational Linguistics: Human Language
  Technologies, Volume 2 (Short Papers)}, pages 15--20, New Orleans, Louisiana.
  Association for Computational Linguistics.

\end{thebibliography}
\bibliographystyle{acl_natbib}

\appendix
\newpage
\onecolumn

\section{List of Text Characteristics}
\label{appendix:metrics}
Tables \ref{fig:metrics_generic} and \ref{fig:metrics_word_level} list all currently implemented metrics along with a short description. For a large number of metrics the Coh-Metrix website\footnote{http://cohmetrix.memphis.edu/cohmetrixhome} provides further details. For word property metrics we also document the source database in Table \ref{fig:metrics_word_level}.

\begin{table*}
\centering\small
\begin{tabular}{lll}
\toprule
 \textbf{Category} & \textbf{Metric Key} & \textbf{Description}  \\ 
 \midrule \midrule
\multirow{12}{*}{Descriptive}
   & DESPC & Number of paragraphs\\
\cmidrule{2-3}
   & DESSC & Number of sentences\\
\cmidrule{2-3}
   & DESWC & Number of words\\
\cmidrule{2-3}
   & DESPL & Average number of sentences per paragraph\\
\cmidrule{2-3}
   & DESPLd & Standard deviation of paragraph lengths (in sentences)\\
\cmidrule{2-3}
   & DESPLw & Average number of words per paragraph\\
\cmidrule{2-3}
   & DESSL & Average number of words per sentence)\\
\cmidrule{2-3}
   & DESSLd & Standard deviation of sentence lengths (in words)\\
\cmidrule{2-3}
   & DESWLsy & Average word length (syllables)\\
\cmidrule{2-3}
   & DESWLsyd & Standard Deviation of word lengths (in syllables)\\
\cmidrule{2-3}
   & DESWLlt & Average word length (letters)\\
\cmidrule{2-3}
   & DESWLltd & Standard Deviation of word lengths (in letters)\\
\midrule
\multirow{4}{*}{Lexical Diversity}
   & LDTTRc & Type-Token Ratio (TTR) computed over content words\\
\cmidrule{2-3}
   & LDTTRa & Type-Token Ratio (TTR) computed over all words\\
\cmidrule{2-3}
   & LDMTLD & Measure of Textual Lexical Diversity (MTLD)\\
\cmidrule{2-3}
   & LDHDD & HD-D lexical diversity index\\
\midrule
\multirow{7}{*}{Syntactic Complexity}
   & SYNLE & Left embeddedness: average words before main verb\\
\cmidrule{2-3}
   & SYNNP & Number of modifiers per noun phrase, mean\\
\cmidrule{2-3}
   & SYNMEDpos & Average edit distance between POS tags of consecutive sentences \\
\cmidrule{2-3}
   & SYNMEDwrd & Average edit distance between consecutive sentences\\
\cmidrule{2-3}
   & SYNMEDlem & Average edit distance between consecutive sentences (lemmatized)\\
\cmidrule{2-3}
   & SYNSTRUTa & Sentence syntax similarity, adjacent sentences, mean\\
\cmidrule{2-3}
   & SYNSTRUTt & Sentence syntax similarity, all combinations, mean\\
\midrule
\multirow{2}{*}{Readability}
   & RDFRE & Flesch Reading Ease \cite{PeterKincaidRobertFishburneJrRichardLRogersBradSChissomDerivationPersonnelCustom}\\
\cmidrule{2-3}
   & READFKGL & Flesch-Kincaid Grade Level \cite{PeterKincaidRobertFishburneJrRichardLRogersBradSChissomDerivationPersonnelCustom}\\

\bottomrule
\end{tabular}
\caption{List of paragraph-level metrics currently supported by TCT}
\label{fig:metrics_generic}
\end{table*}

\begin{table*}
\centering\small
\begin{tabular}{lll}
\toprule
 \textbf{Category} & \textbf{Metric Key} & \textbf{Description}  \\ 
 \midrule

\midrule
\multirow{20}{*}{Incidence Scores}
   & TOKEN\_ATTRIBUTE\_RATIO\_ALHPA & Alphanumerical tokens\\
\cmidrule{2-3}
   & TOKEN\_ATTRIBUTE\_RATIO\_DIGIT & Tokens consisting of digits\\
\cmidrule{2-3}
   & TOKEN\_ATTRIBUTE\_RATIO\_PUNCT & Punctuation tokens\\
\cmidrule{2-3}
   & TOKEN\_ATTRIBUTE\_RATIO\_URL & URLs\\
\cmidrule{2-3}
   & TOKEN\_ATTRIBUTE\_RATIO\_EMAIL & E-mail addresses\\
\cmidrule{2-3}
   & WORD\_SET\_INCIDENCE\_WRDPRP1s & First person singular pronouns\\
\cmidrule{2-3}
   & WORD\_SET\_INCIDENCE\_WRDPRP1p & First person plural pronouns\\
\cmidrule{2-3}
   & WORD\_SET\_INCIDENCE\_WRDPRP2 & Second person pronouns\\
\cmidrule{2-3}
   & WORD\_SET\_INCIDENCE\_WRDPRP3s & Third person singular pronouns\\
\cmidrule{2-3}
   & WORD\_SET\_INCIDENCE\_WRDPRP3p & Third person plural pronouns\\
\cmidrule{2-3}
   & WORD\_SET\_INCIDENCE\_CNCCaus & Causal connectives\\
\cmidrule{2-3}
   & WORD\_SET\_INCIDENCE\_CNCLogic & Logical connectives\\
\cmidrule{2-3}
   & WORD\_SET\_INCIDENCE\_CNCTemp & Temporal connectives\\
\cmidrule{2-3}
   & WORD\_SET\_INCIDENCE\_CNCAdd & Additive connectives\\
\cmidrule{2-3}
   & WORD\_SET\_INCIDENCE\_CNCPos & Positive connectives\\
\cmidrule{2-3}
   & WORD\_SET\_INCIDENCE\_CNCNeg & Negative connectives\\
\cmidrule{2-3}
   & WORD\_PROPERTY\_WRDNOUN & Incidence score for POS tag {'PROPN', 'NOUN'}\\
\cmidrule{2-3}
   & WORD\_PROPERTY\_WRDVERB & Incidence score for POS tag {'VERB'}\\
\cmidrule{2-3}
   & WORD\_PROPERTY\_WRDADJ & Incidence score for POS tag {'ADJ'}\\
\cmidrule{2-3}
   & WORD\_PROPERTY\_WRDADV & Incidence score for POS tag {'ADV'}\\
\midrule
\multirow{16}{*}{Word Property}
   & WORD\_PROPERTY\_WRDFRQc & Mean Word frequency\textsuperscript{*}, content words\\
\cmidrule{2-3}
   & WORD\_PROPERTY\_WRDFRQa & Mean Word frequency\textsuperscript{*}, all words\\
\cmidrule{2-3}
   & WORD\_PROPERTY\_WRDFRQmc & Min Word frequency\textsuperscript{*}\\
\cmidrule{2-3}
   & WORD\_PROPERTY\_WRDFAMc & Mean Familiarity\textsuperscript{+}, content words only\\
\cmidrule{2-3}
   & WORD\_PROPERTY\_WRDCNCc & Mean Concreteness\textsuperscript{+}, content words only\\
\cmidrule{2-3}
   & WORD\_PROPERTY\_WRDIMGc & Mean Imagability\textsuperscript{+}, content words only\\
\cmidrule{2-3}
   & WORD\_PROPERTY\_WRDMEAc & Mean Meaningfulness\textsuperscript{+}\\
\cmidrule{2-3}
   & WORD\_PROPERTY\_WRDPOLc & Mean Polysemy\textsuperscript{\textdagger}\\
\cmidrule{2-3}
   & WORD\_PROPERTY\_WRDHYPn & Mean Hypernymy\textsuperscript{\textdagger} (nouns)\\
\cmidrule{2-3}
   & WORD\_PROPERTY\_WRDHYPv & Mean Hypernymy\textsuperscript{\textdagger} (verbs)\\
\cmidrule{2-3}
   & WORD\_PROPERTY\_WRDHYPnv & Mean Hypernymy\textsuperscript{\textdagger} (verbs and nouns)\\
\cmidrule{2-3}
   & WORD\_PROPERTY\_AOA & Mean Age of Acqusition \cite{KupermanAge-of-acquisitionWords}\\
\cmidrule{2-3}
   & WORD\_PROPERTY\_AOA\_MAX & Max Age of Acqusition \cite{KupermanAge-of-acquisitionWords}\\
\cmidrule{2-3}
   & WORD\_PROPERTY\_CONCRETENESS & Mean Concreteness \cite{Brysbaert2014ConcretenessLemmas}\\
\cmidrule{2-3}
   & WORD\_PROPERTY\_PREVALENCE & Mean Prevalence \cite{BrysbaertWordLemmas}\\
\cmidrule{2-3}
   & WORD\_PROPERTY\_PREVALENCE\_MIN & Minimum Prevalence \cite{BrysbaertWordLemmas}\\

\bottomrule
\end{tabular}
\caption{List of word-level metrics currently supported by TCT. Common underlying word databases: \\ \textsuperscript{*}SpaCy \cite{Honnibal_spaCy_Industrial-strength_Natural_2020} \textsuperscript{+}MRC \cite{Coltheart2018TheDatabase:} \textsuperscript{\textdagger} WordNet \cite{Fellbaum2010WordNet}}
\label{fig:metrics_word_level}
\vspace{-0.5cm}
\end{table*}

\newpage
\section{Sample Use Cases}
\label{section:use_cases}

In this section we demonstrate how users can gain actionable insights on existing evaluation data using TCT, with minimal amount of additional work. 
The examples provided below can be reproduced by Python notebooks we included in the examples folder of the TCT repository.


\subsection{Predicting Accuracy of OPT Baselines}
\label{section:predict_model_perf}

Despite the recent success of large pre-trained language models (LLM), there are still ongoing debates regarding how good they really are, and how to evaluate that. After all, LLMS such as PaLM \citep{chowdhery2022palm} or DeBERTa \citep{he2020deberta} have saturated the performance on benchmarks, even outperforming human scores at times,
but, at the same time, there still exist a myriad of seemingly trivial scenarios in which they fail.

\paragraph{Experimental setup} 
In this demonstration we offer an alternative approach: We take existing data from the evaluation of the 6.7B OPT baseline \cite{Zhang2022OPT:Models} then attempt to use simple data characteristics to identify interpretable subsets of the dataset on which OPT's performance substantially differs from its overall high accuracy. We use the HellaSwag task \cite{Zellers2019HellaSwag:Sentence},\footnote{We chose the HellaSwag task for this demo as it had sufficiently many examples in the test set and showed the most interesting correlations out of all tasks the model was evaluated on prior.} a common-sense inference task that is trivial to solve for humans but challenging for LLMs.

To evaluate OPT models on this task, prompts corresponding to different choices were scored with the LLMs and the answer with the lowest perplexity was considered to be the choice of the model. For each data point in the test set, we consider two text fragments: the prompt corresponding to the correct answer and the concatenation of all the prompts corresponding to incorrect answers (see Table \ref{tab:text_features} for an example).
With a single command using the \texttt{command\_line\_tool.py} we compute characteristics for the extracted texts and load the results into a notebook. We also load the result of the model evaluation, which is a single binary variable per data point describing whether the model predicted the right answer.

\begin{figure*}
    \centering
    \begin{subfigure}[b]{0.3\linewidth}
        \includegraphics[width=\linewidth]{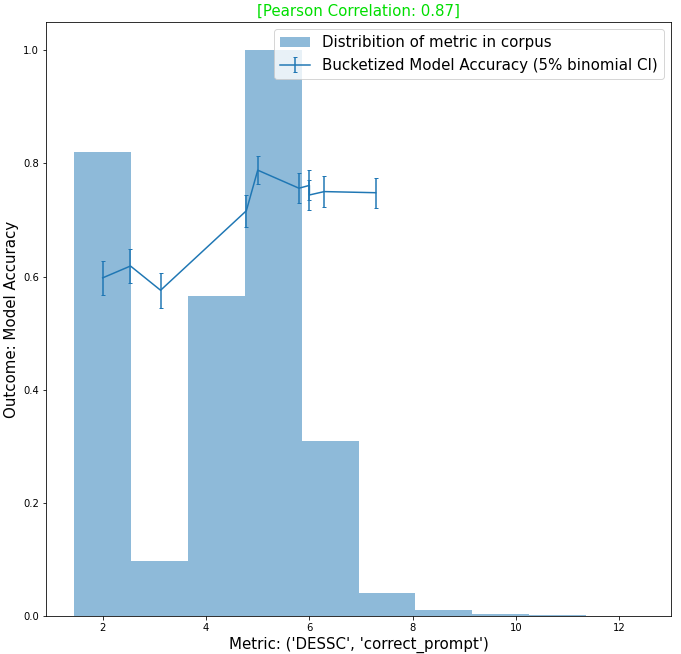}
        \caption{Sentence Count}
        \label{fig:a}
    \end{subfigure}
    \begin{subfigure}[b]{0.3\linewidth}
        \includegraphics[width=\linewidth]{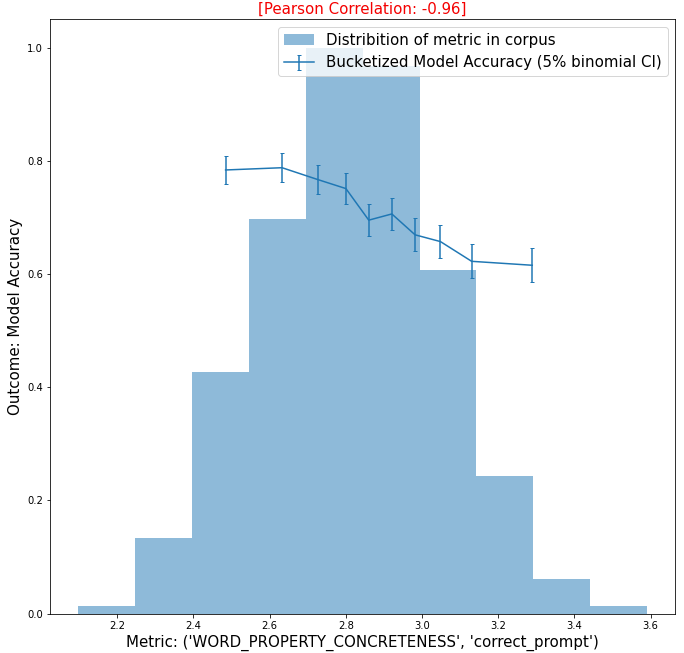}
        \caption{Concreteness}
        \label{fig:b}
    \end{subfigure}
    \begin{subfigure}[b]{0.3\linewidth}
        \includegraphics[width=\linewidth]{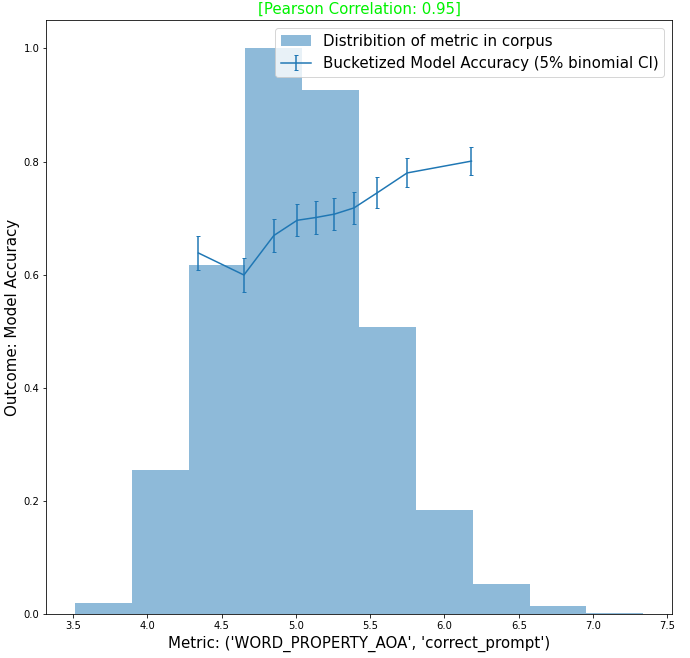}
        \caption{Age of Acquisition}
        \label{fig:c}
    \end{subfigure}
    \caption{
        Correlations between text characteristics and accuracy for OPT experiments,
    }
    \label{fig:individual_correlations}
\end{figure*}

\paragraph{Results} First, we inspect correlations between individual metrics and model performance. This analysis tool orders data points with respect to a particular TCT metric, groups them into buckets of 100 data points and computes model accuracy for each bucket. We  find several different data characteristics that show high correlation with model performance, for example number of sentences per prompt or average concreteness \citep{Brysbaert2014ConcretenessLemmas} of words. A visualisation of these results is shown in Figure \ref{fig:individual_correlations}.

\begin{figure*}[t]
    \centering
    \includegraphics[width=0.9\linewidth]{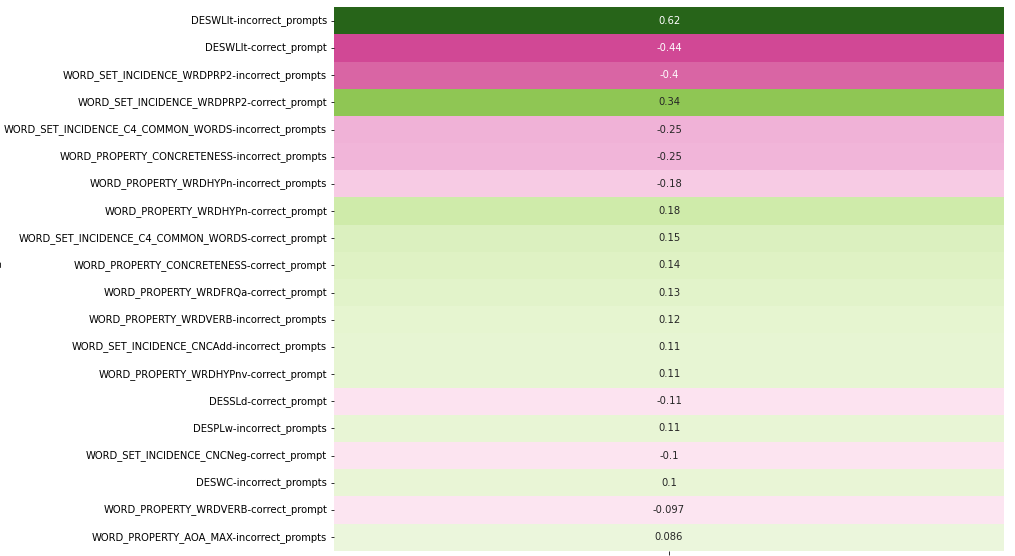}
    \caption{Coefficients of logistic regression visualised by TCT for OPT experiment.}
    \label{fig:coeffs}
\end{figure*}

\begin{figure}
  \centering
  \includegraphics[height=55mm]{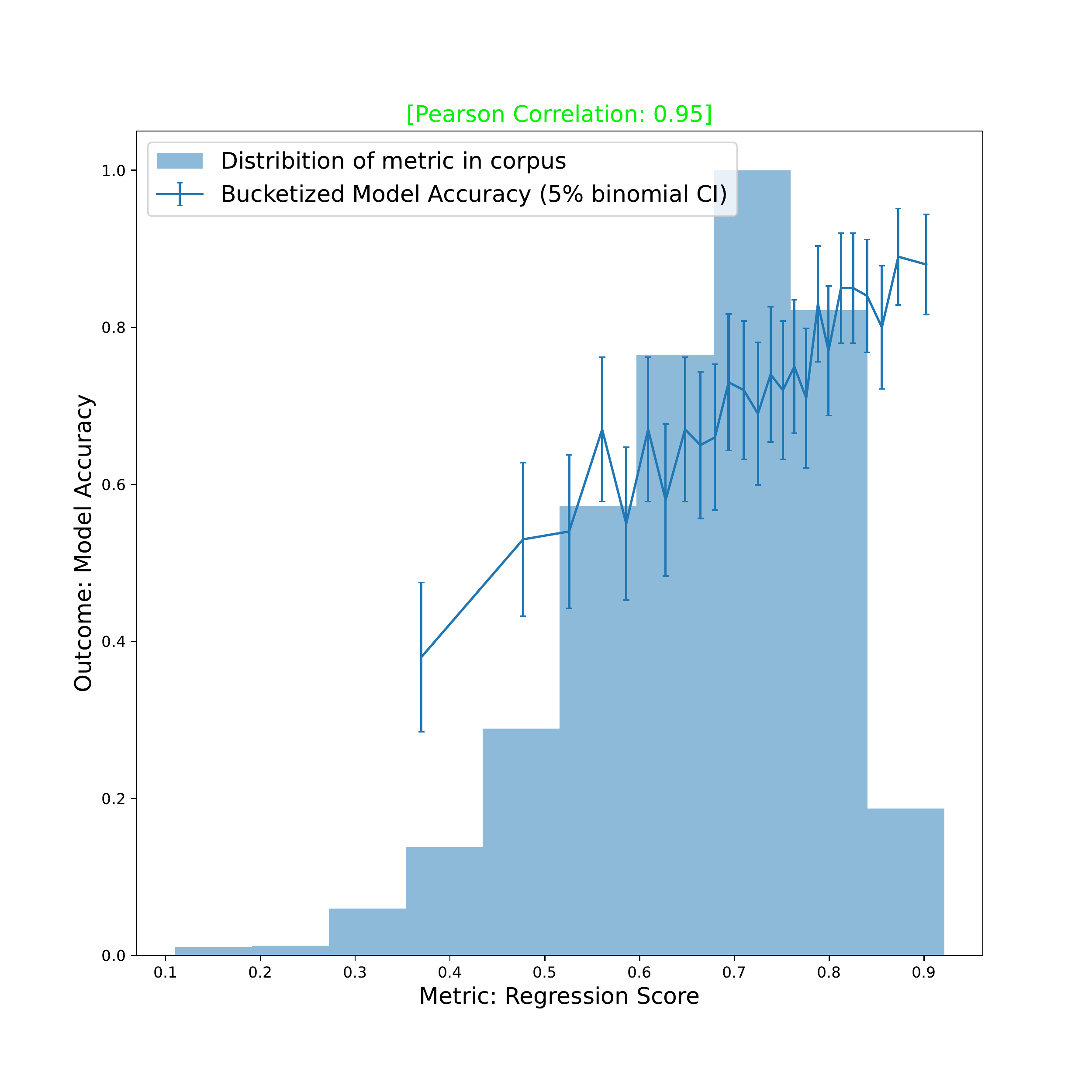}
  \caption{The TCT analysis tool visualizing the regression model applied to held out data in the OPT analysis.}
  \label{fit:regression_fit}
  \vspace{-0.1cm}
\end{figure}

Secondly, we employ the TCT class named \texttt{PredictFromCharacteristicsAnalysis} to fit a logistic regression model using all characteristics to predict whether the model will yield a correct answer for a particular data point. The tool computes the regression scores on a held out part of the dataset and visualizes model accuracy with respect to this score per data bucket, as shown in Figure~\ref{fit:regression_fit}.
We find more variance between the best and worst performing buckets compared to the single variable analysis.
On the bucket with the highest predicted score the OPT baselines yield a 0.9 accuracy, but in the lowest scoring bucket the accuracy is below 0.4, which approaches the random baseline of 0.25. 
To interpret the fitted regression model, we inspect its coefficients,\footnote{Since inputs to the regression were scaled to unit variance, direct comparison of coefficients is meaningful} illustrated in Figure~\ref{fig:coeffs}. 
Interestingly, coefficients for given characteristics often yield opposite signs associated with the correct and incorrect answers, indicating that they are in fact, on their own, predictive of the correctness of an answer.
For instance, the \textit{DESWLlt} metric (mean number of letters per word) has coefficients of -0.44 and 0.62 for the \texttt{correct\_prompt} and \texttt{incorrect\_prompts} features, respectively. 

We argue that such analyses are useful from two perspectives: i) Analyses that uncover patterns in what characteristics make examples difficult help us improve our understanding of how well a model has in fact learned the task we intended it to.
This, in turn, provides a better estimate of the wider applicability of a model.
ii) If one knows which text characteristics lead to poor performance from LLMs, one could improve the dataset's coverage for characteristics associated with low model performance -- e.g. one could curate data points including tokens with low concreteness scores.




Table \ref{tab:text_features}, Figure \ref{fig:individual_correlations} and Figure \ref{fig:coeffs} illustrate the model analysis process described previously in this section.

\begin{figure*}[t]
  \centering
  \renewcommand{\arraystretch}{3}
  \begin{tabularx}{\textwidth}{ll}
    \textbf{Correct Prompt} & Roof shingle removal: A man is sitting on a roof. He starts pulling up roofing on a roof. \\
    \hline
    \textbf{Incorrect Prompts} & \makecell[l]{
        Roof shingle removal: A man is sitting on a roof. He is using wrap to wrap a pair of skis. \\
        Roof shingle removal: A man is sitting on a roof. He is ripping level tiles off. \\
        Roof shingle removal: A man is sitting on a roof. He is holding a rubik's cube.
    }
  \end{tabularx}
  \caption{Example of text features extracted from the HellaSwag evaluation of the OPT model}
  \label{tab:text_features}
\end{figure*}

\subsection{Gender Bias in Co-reference Resolution Models}

Second, we would like to illustrate how TCT can aid in identifying biases in NLP systems, by revealing gender bias in coreference resolution systems.

\paragraph{Experimental Setup}
We use a coreference resolution model proposed by ~\citet{lee-etal-2017-end} and the WinoBias dataset \cite{zhao-etal-2018-gender}. The model is evaluated using exact match to compute accuracy.
To capture gender statistics, we configure a new Word Property metric ``genderedness'' based on Labor Force Statistics\footnote{\url{https://github.com/uclanlp/corefBias}} and compute it on two text fragments (the two spans of the ground truth co-reference). A higher genderedness score represents that the occupation is associated with a female stereotype and vice versa. For pronominal references, we assign 100 to female ones (e.g. ``she'', ``her'') and 0 to male ones (e.g. ``he'', ``his''). We add the difference between the two characteristics as an additional feature for analysis.

\begin{figure}[!htbp]
    \centering
  \includegraphics[height=60mm]{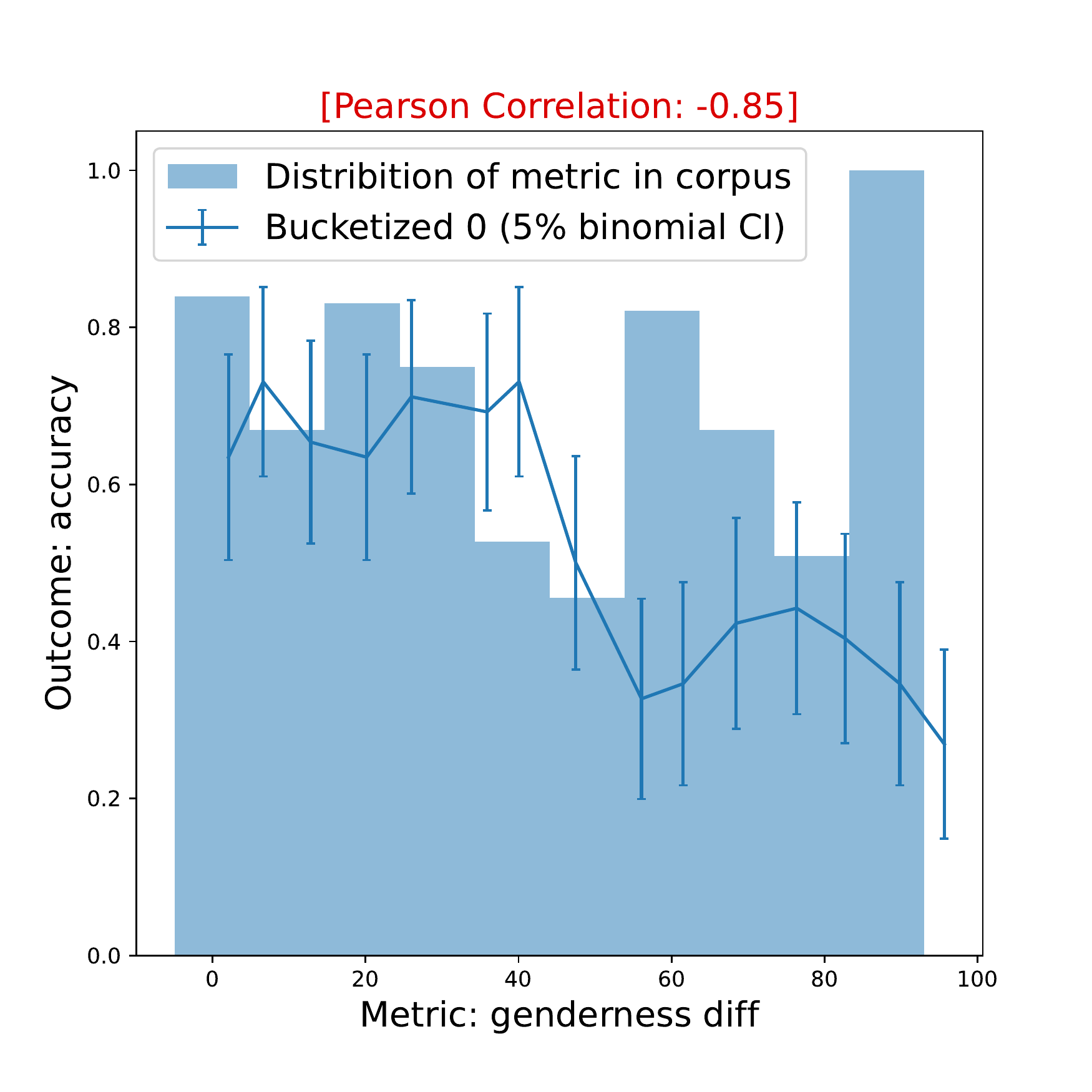}
  \captionof{figure}{Genderness difference hurts the performance of a coreference resolution model.}
  \label{fit:coref_bias}
  \vspace{-0.1cm}
\end{figure}

\paragraph{Results}
The analysis obtained by the TCT toolkit is illustrated in Figure~\ref{fit:coref_bias}. There is a negative correlation between model accuracy and the genderness difference between the occupation and the pronominal reference. In other words, if a female stereotypical occupation and a male pronoun co-occur in a test example (e.g. ``nurse'' and ``he'') or a male stereotypical occupation and a female pronoun (e.g. ``constructor'' and ``she'') co-occurs, the model is more likely to make a wrong prediction.



\subsection{NLLB: Interpretable Fluctuations of Translation Performance}

\begin{figure}
    \centering
    \includegraphics[width=.5\textwidth]{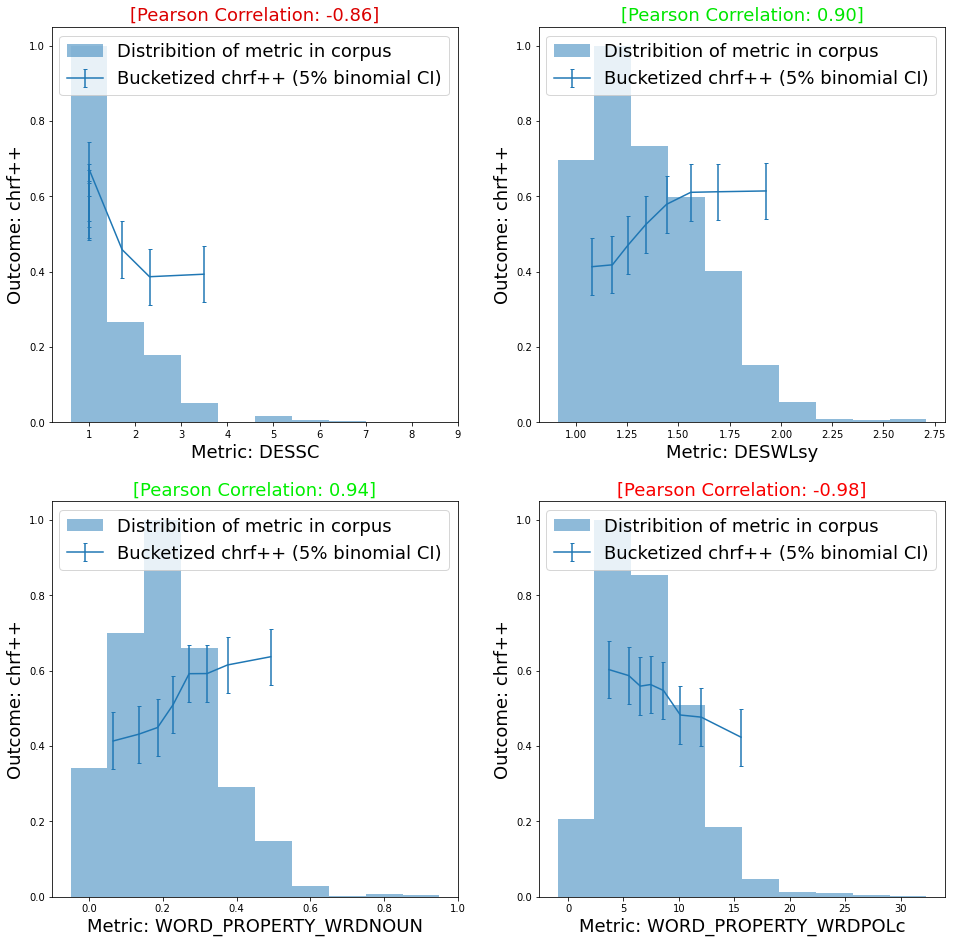}
    \caption{Correlations between text characteristics and chrf++ scores in the NLLB analysis.}
    \label{fig:het_nllb}
     \vspace{-0.1cm}
\end{figure}

A third example of a task that could benefit from using TCT in analyses is \textit{Neural Machine Translation} (NMT). We apply TCT to source sentences to identify patterns in translation success for an off-the-shelf NMT system.

\paragraph{Experimental Setup} To investigate performance heterogeneity in translation models, we use the No Language Left Behind 1.3B distilled model and the English-Russian validation split of the multi-domain dataset from the same work~\citep{costa2022no}. We use the HuggingFace transformers translation pipeline for easy inference~\citep{wolf2019huggingface}. We extract translations using the pipeline, and employ the \texttt{chrf++} metric to measure success per individual data point~\citep{popovic2017chrf++}.\footnote{Note: we use this as it has better per data point properties than other corpus statistics like BLEU~\citep{papineni2002bleu}.} Using the toolkit we characterize the English source data with default settings.

\paragraph{Results}
Surprisingly, we find significant heterogeneity as seen in Figure~\ref{fig:het_nllb} and below. In particular, more sentences, more verbs, polysemy for content words, and chat-like messages lead to performance drops. Conversely, more nouns and words with more syllables correlate with better \texttt{chrf++} scores.

\begin{figure*}
    \centering
    \includegraphics[width=.7\textwidth]{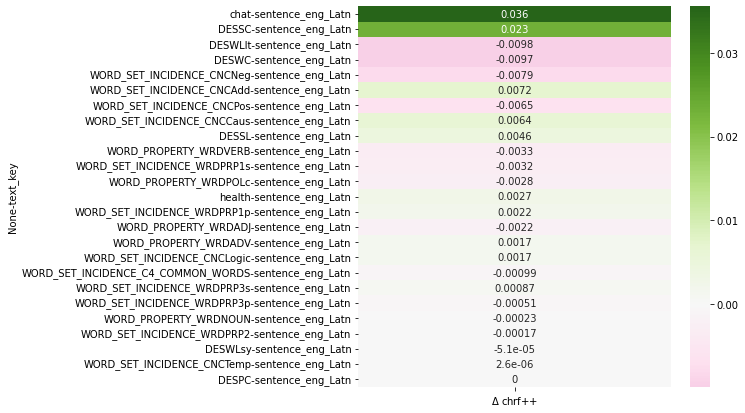}
    \includegraphics[width=\textwidth]{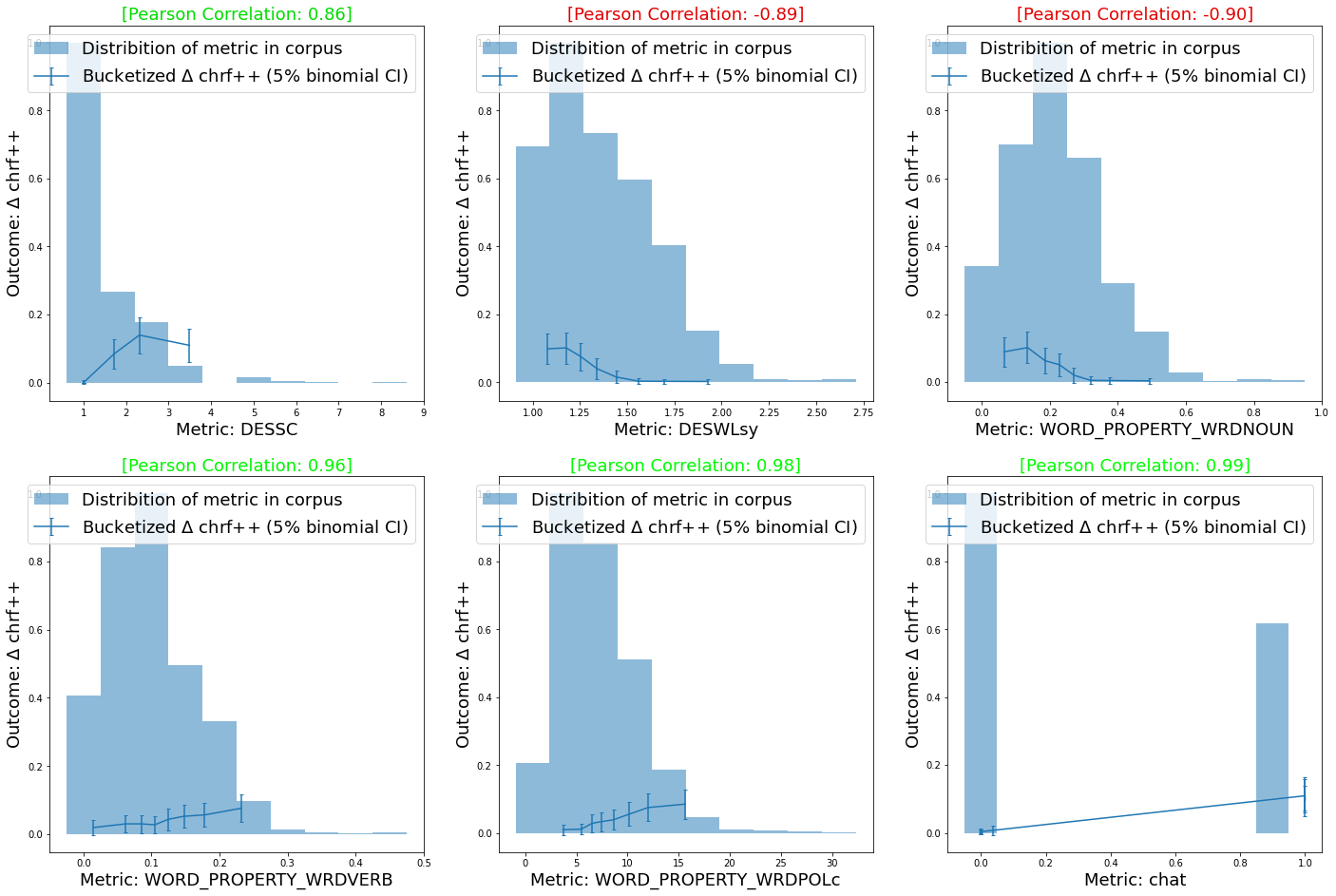}
    \caption{Top: A regression on the per-sentence treatment effect between a translation run through NLLB with and without sentence segmentation. Positive values indicate correlation with improved chrf++ from segmentation over the base model. Bottom: As can be seen there is even heterogeneity in treatment effects. Sentence splitting has the most positive effect correlation with the chat corpus and for the evaluation points with many sentences.}
    \label{fig:nllb_regression_treatment_effect}
\end{figure*}

The driver of this heterogeneity may be deceptive. The HuggingFace translation pipeline does not keep track of the underlying model's training distribution. It would not know that the NLLB model was trained on sentence pairs and the evaluation data contains multi-sentence datapoints. An appropriate way to match the training distribution would instead be to split by sentences and translate individual sentences before re-concatenating. In fact, if we take this approach, we find that performance levels out with the biggest improvements coming from the largest sources of heterogeneity (Figure~\ref{fig:nllb_regression_treatment_effect}). 
This demonstration shows the power of TCT for debugging model workflows. With many layers of abstraction, it is easy to forget that underlying models are likely trained on a particular data distribution.

\paragraph{Additional Details}
\label{appendix:nllb_more}

\begin{figure*}
    \centering
    \includegraphics[width=.7\textwidth]{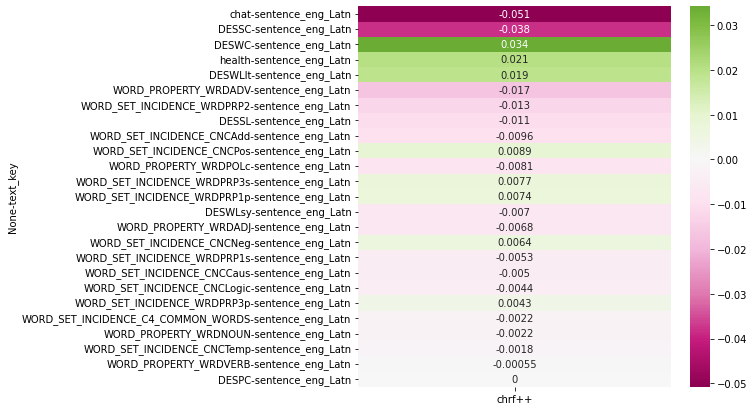}
    \includegraphics[width=\textwidth]{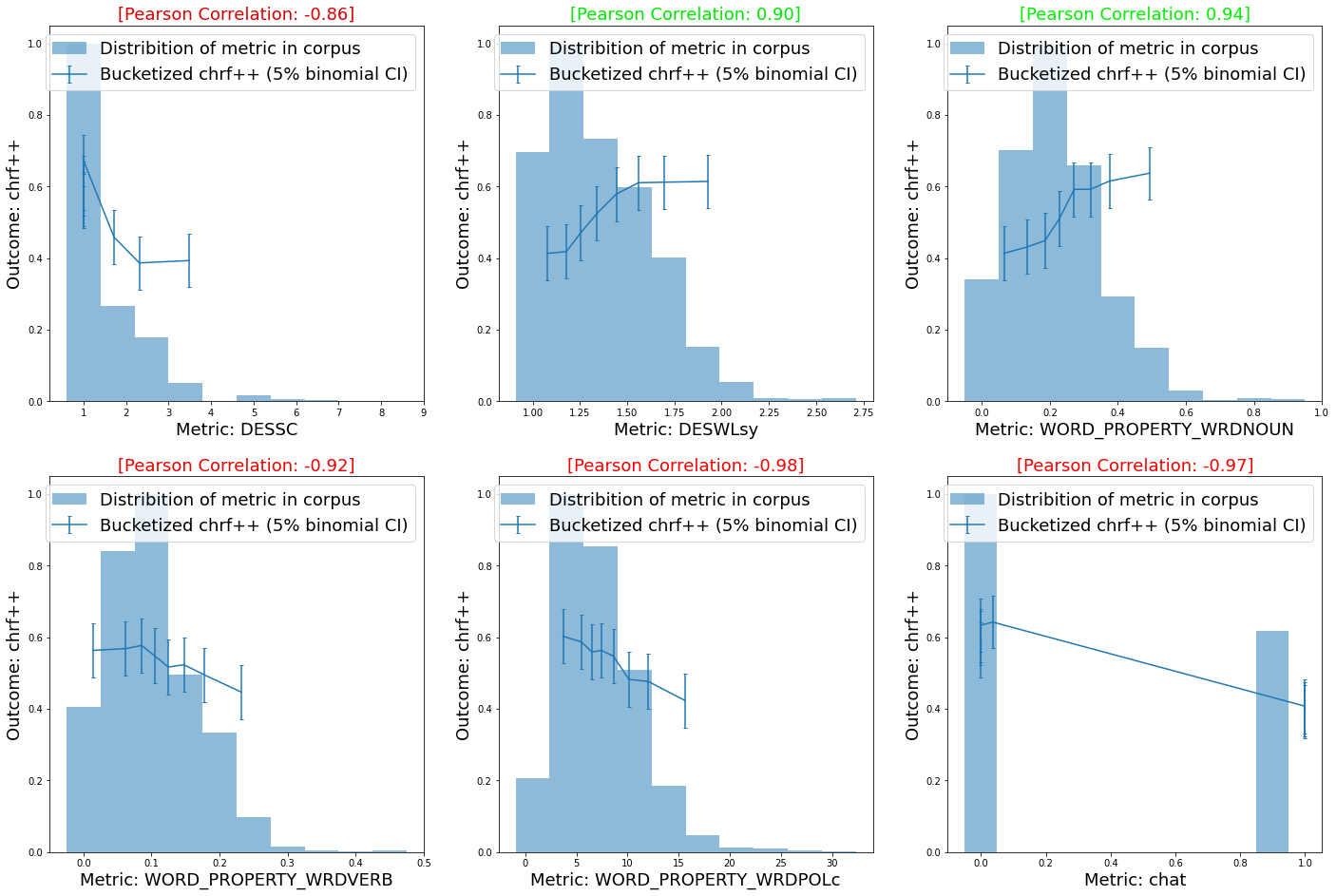}
    \caption{Top: A regression on the chrf++ score of a model pipeline using NLLB with no segmentation. Positive values indicate improved score, lower values indicate a negative correlation with score. Bottom: As can be seen there is even heterogeneity in treatment effects for several data characteristics.}
    \label{fig:nosegment_app}
\end{figure*}

\begin{figure*}
    \centering
    \includegraphics[width=.7\textwidth]{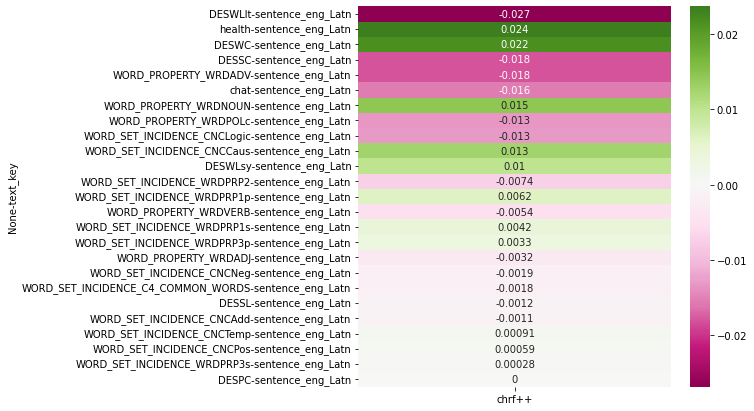}
    \includegraphics[width=\textwidth]{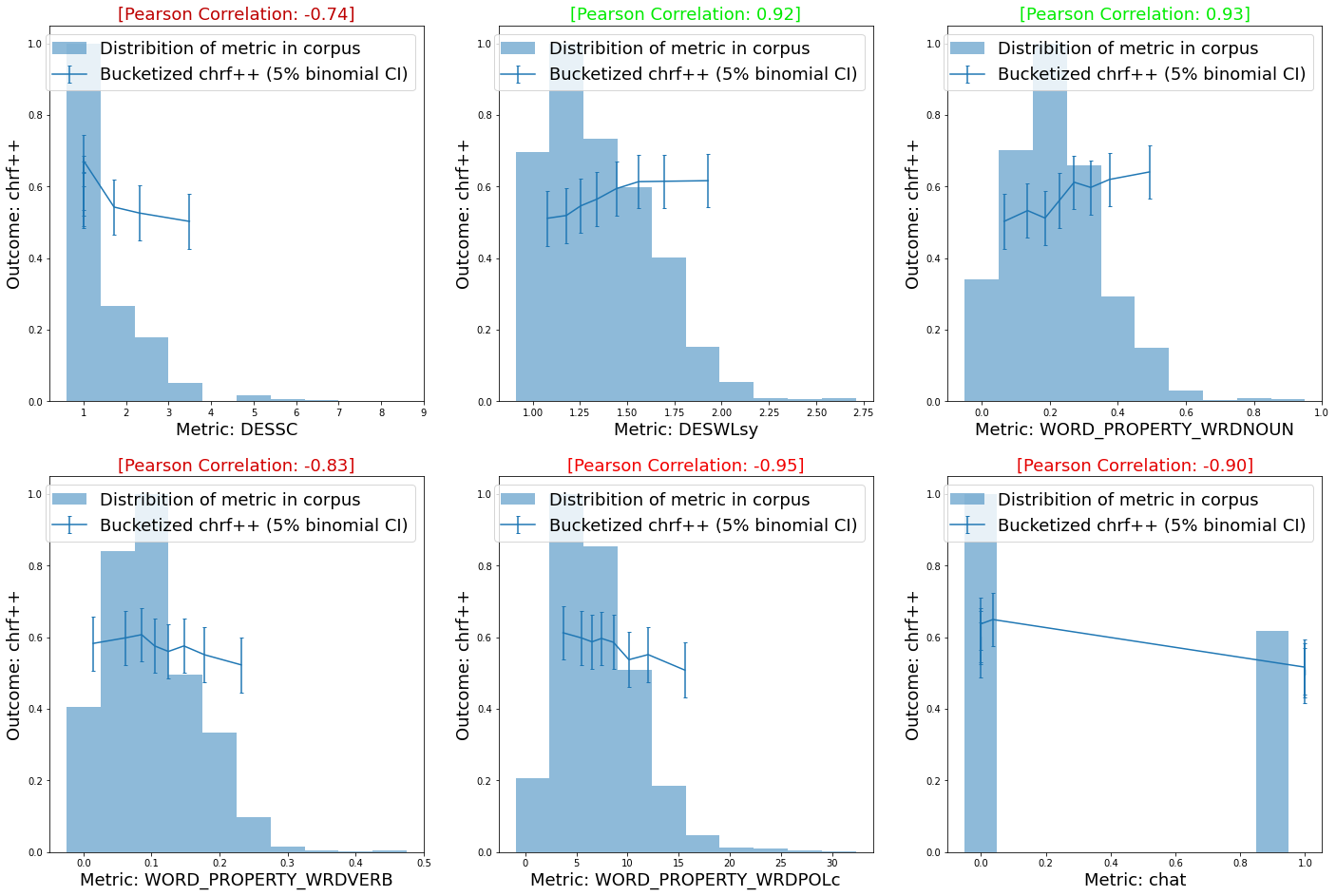}
    \caption{Top: A regression on the chrf++ score of a model pipeline using NLLB with segmentation. Positive values indicate improved score, lower values indicate a negative correlation with score. Bottom: As can be seen there is still some heterogeneity in treatment effects for several data characteristics but they have significantly flattened.}
    \label{fig:segment_app}
\end{figure*}

Figure~\ref{fig:nosegment_app} shows the distribution of data-characterized performance for NLLB using the HuggingFace translate tool with no modification (other than increasing the maximum generated length to 512 tokens). Figure~\ref{fig:segment_app} shows the distribution of chrf++ scores for NLLB with sentence segmentation\footnote{Sentence segmentation by SpaCy~\citep{Honnibal_spaCy_Industrial-strength_Natural_2020}}. We pass each sentence in a batch to the segmentation pipeline before re-concatenating them by adding a space between sentences (since we only use English and Russian for this demonstration this is an appropriate concatenation method). Finally, Figure~\ref{fig:nllb_regression_treatment_effect} shows the treatment effect. For each sentence we subtract the segmented NLLB chrf++ score from the unsegmented chrf++ score. Then we run the TCT toolkit on this outcome measure. We show that performance increases are such that they level out performance heterogeneity to some extent. 

In Table~\ref{tab:nllb_examples_multisentence} we demonstrate how the unsegmented NLLB model can drop out entire portions of the translation in multi-sentence validation datapoints. This is likely what leads to performance drops. The segmented version fixes this. As such, we suggest that TCT should be run at eval time even when using a known model that has been validated in the past. Different environmental setups can lead to failure modes such as this one that can be difficult to detect without data characterization.

\begin{table*}{}\fontencoding{T2A}\selectfont
    \centering
    \begin{tabular}{l|p{12cm}}
         English & Yeah that would be so fun! {\color{red} It's really easy honestly, there's a bit of skill with steering} but once you get the hang of it it feels super natural.\\
         True & Да, было бы очень весело! {\color{red}Честно говоря, это очень просто, нужен небольшой навык руления,} но как только вы поймете, ощущение будет очень естественным.\\
        NLLB & Да, это было бы так весело! но как только ты научишься управлять, это будет очень естественно.\\
        NLLB (seg) &Да, это было бы так весело! {\color{red}Это очень просто, честно говоря, требуется немного мастерства,} но как только ты научишься управлять, это будет очень естественно.\\
         \hline
        English & {\color{red} That's right.} How  is your family? {\color{red}how many of you are there?}\\
        True & {\color{red}Точно.} Как твоя семья? {\color{red}Сколько вас?}\\
        NLLB & Как ваша семья?\\
    NLLB (seg) &{\color{red} - Да, это так.} Как твоя семья?{ \color{red} Сколько вас там?} \\
    \end{tabular}
    \caption{NLLB without proper segmentation misses entire portions of the context. For example, the red-highlighted portion is missing from the non-segmented NLLB text, but present elsewhere.}
    \label{tab:nllb_examples_multisentence}
\end{table*}


\end{document}